%% file: egpaper_final.tex
\documentclass[10pt,twocolumn,letterpaper]{article}
\pdfoutput=1
\usepackage{iccv}
\usepackage{times}
\usepackage{epsfig}
\usepackage{graphicx}
\usepackage{amsmath}
\usepackage{amssymb}
\usepackage{subcaption}
\usepackage{overpic}
\usepackage{multicol}
\usepackage{multirow}
\usepackage{bm}
\usepackage{balance}
\usepackage{booktabs}
\usepackage[accsupp]{axessibility}
\usepackage[pagebackref=true,breaklinks=true,letterpaper=true,colorlinks,bookmarks=false]{hyperref}

\iccvfinalcopy % *** Uncomment this line for the final submission

\def\iccvPaperID{5852} % *** Enter the ICCV Paper ID here

\ificcvfinal\fi

\begin{document}

%%%%%%%%% TITLE
\title{DREAM: Efficient Dataset Distillation by Representative Matching}

\author{
Yanqing Liu \textsuperscript{1,2}\thanks{Equal contribution.} 
\quad Jianyang Gu\textsuperscript{1,2}\footnotemark[1]
\quad Kai Wang\textsuperscript{1}\thanks{Project lead.}
\quad Zheng Zhu\textsuperscript{3}
\quad Wei Jiang\textsuperscript{2}
\quad Yang You\textsuperscript{1}\thanks{Corresponding author.}
% \thanks{Corresponding author.}
% 
% \\
% 
% 
% 
\\
\textsuperscript{1}{National University of Singapore}
\quad \textsuperscript{2}{Zhejiang University}
\quad \textsuperscript{3}{Tsinghua University}
\\
\small{\texttt{\{yanqing\_liu, gu\_jianyang,  jiangwei\_zju\}@zju.edu.cn}} \\
\small{\texttt{\{kai.wang, youy\}@comp.nus.edu.sg}}
\quad \small{\texttt{zhengzhu@ieee.org}}
\\
\small{Code: \url{https://github.com/lyq312318224/DREAM}}
}

\maketitle
\ificcvfinal\thispagestyle{empty}\fi

%%%%%%%%% ABSTRACT
\begin{abstract}
   Dataset distillation aims to synthesize small datasets with little information loss from original large-scale ones for reducing storage and training costs. Recent state-of-the-art methods mainly constrain the sample synthesis process by matching synthetic images and the original ones regarding gradients, embedding distributions, or training trajectories. Although there are various matching objectives, currently the strategy for selecting original images is limited to naive random sampling. 
   We argue that random sampling overlooks the evenness of the selected sample distribution, which may result in noisy or biased matching targets.
   Besides, the sample diversity is also not constrained by random sampling. These factors together lead to optimization instability in the distilling process and degrade the training efficiency. Accordingly, we propose a novel matching strategy named as \textbf{D}ataset distillation by \textbf{RE}present\textbf{A}tive \textbf{M}atching (DREAM), where only representative original images are selected for matching. DREAM is able to be easily plugged into popular dataset distillation frameworks and reduce the distilling iterations by more than 8 times without performance drop. Given sufficient training time, DREAM further provides significant improvements and achieves state-of-the-art performances.
\end{abstract}

%%%%%%%%% BODY TEXT
\input{sections/intro}
\input{sections/related}

\input{sections/method}
\input{sections/experiment}

\section{Conclusion}
In this paper, we propose a novel Dataset distillation by REpresentAtive Matching (DREAM) strategy to address the training efficiency problem for dataset distillation. 
By only matching with the representative original images, DREAM reduces the optimization instability, and reaches a smoother and more robust training process. 
It is able to be easily plugged into popular dataset distillation frameworks to reduce the training iterations by more than 8 times without performance drop. 
The stable optimization also provides higher final performance and generalization capability. 
The more efficient matching allows future works to design more complicated matching metrics. 

\section{Limitations and Future Works}
Although the proposed DREAM strategy significantly improves the training efficiency of optimization-based dataset distillation methods, the calculation burden is still large when the image size and the class number increases. 
It is difficult for these methods to handle ultra large-scale datasets like ImageNet~\cite{deng2009imagenet}, even if the training efficiency has been improved by DREAM. 
We will explore more resource-friendly ways to conduct dataset distillation in future works. 

\section*{Acknowledgement}
This research is supported by the National Research Foundation, Singapore under its AI Singapore Programme (AISG Award No: AISG2-PhD-2021-08-
008). Yang You's research group is being sponsored by NUS startup grant (Presidential Young Professorship), Singapore MOE Tier-1 grant, ByteDance grant, ARCTIC grant, SMI grant and Alibaba grant.
The research is also supported by the National Natural Science Foundation of China (No. 62173302).
{\small
\bibliographystyle{ieee_fullname}
\bibliography{egbib}
}

\input{sections/supp}

\end{document}

%% file: sections/intro.tex
\section{Introduction}
\label{sec:intro}
Deep learning has made remarkable achievements in the computer vision society~\cite{he2016deep,dosovitskiy2020image,redmon2016you,liu2021swin,tsai2018learning,goodfellow2020generative,danelljan2017eco,zheng2023preventing}, and the success is closely related to a large amount of efforts in data collection and annotation. 
But along with the progress of these efforts, the huge amount of data, in turn, becomes a barrier to both storage and training~\cite{zhao2020dc,he2022masked}. 
Many methods are introduced to reduce the scale of datasets~\cite{wang2018dataset,sorscher2022beyond,qin2023infobatch,zhou2023dataset}.
Among these, dataset distillation, aiming at condensing large-scale datasets into smaller ones with little information loss, has become a hot topic to tackle the problem of data burden~\cite{mtt,wang2022cafe,idc,cui2022dc,ftd}. 

\begin{figure}
\centering
\begin{subfigure}{0.48\textwidth}
\includegraphics[width=\textwidth]{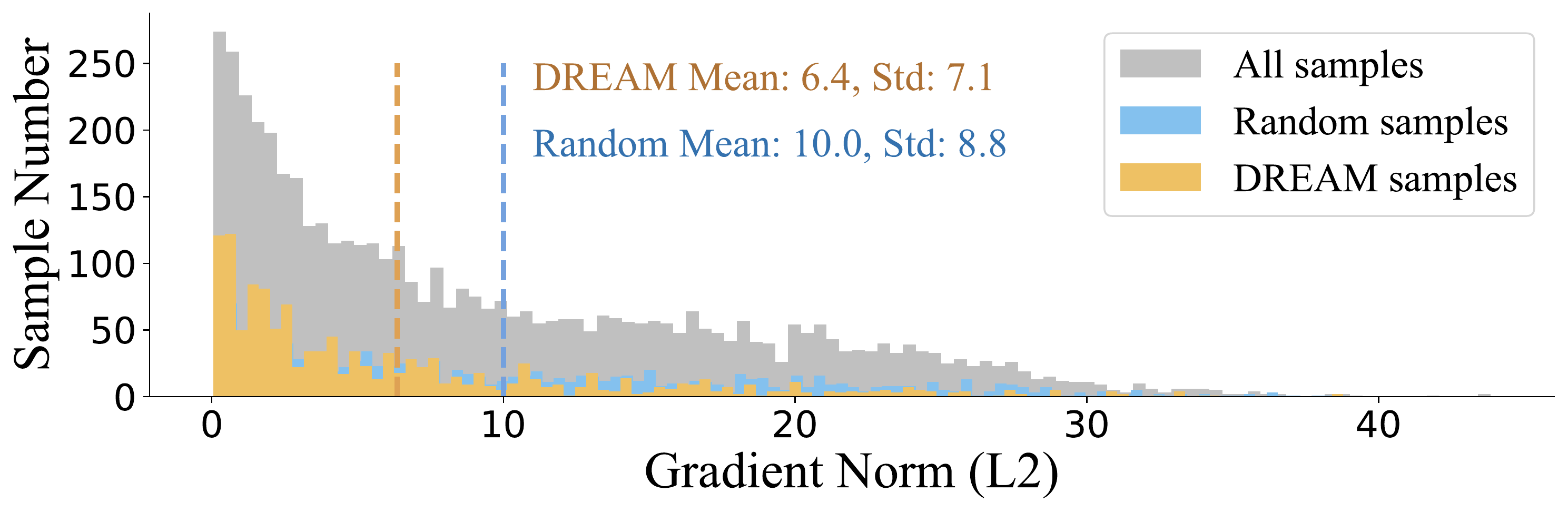}
\caption{The gradient norm distribution of the plane class in CIFAR10. }
\label{fig:gradient}
\end{subfigure}
\begin{subfigure}{0.5\textwidth}
    \includegraphics[width=\textwidth]{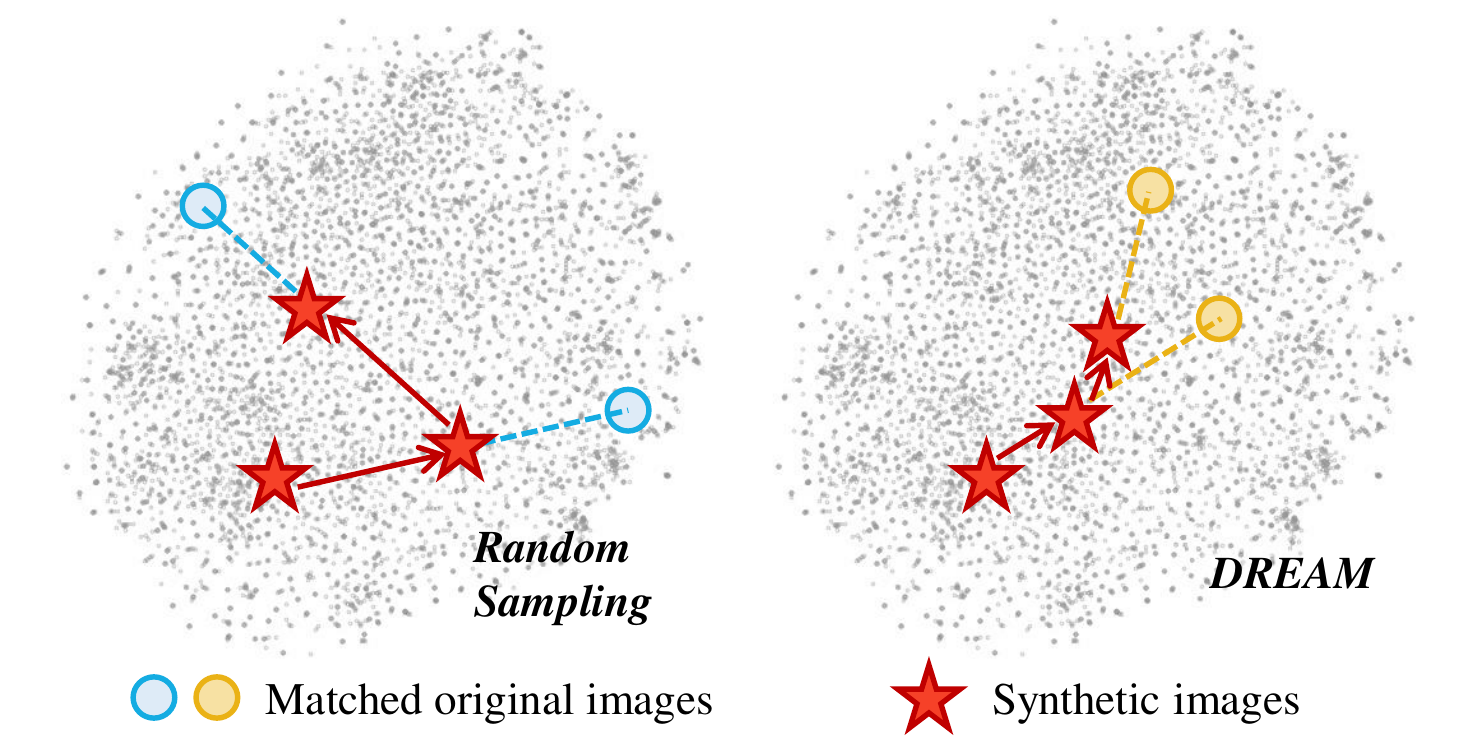}
    \caption{The migration of synthetic samples during training. }
    \label{fig:intro}
\end{subfigure}
\caption{Samples on the decision boundaries usually provide larger gradients, which biases the gradient matching optimization. Random sampling (left) overlooks the evenness of of the selected sample distribution, resulting in unstable optimization process of the synthesized samples. By only matching with proper gradients from representative original samples, our proposed DREAM (right) greatly improves the training efficiency of dataset distillation tasks. Best viewed in color. }
\end{figure}

Dataset distillation methods are roughly divided into two categories: coreset-based and optimization-based. 
Coreset-based method employ certain metrics to heuristically select samples for representing the original dataset~\cite{lapedriza2013all,toneva2018empirical}. 
However, it is difficult to rely on a small proportion of original samples to contain the information of the whole dataset, resulting in low compression rate. 
Optimization-based methods alleviate the defect by incorporating image synthesis to introduce more information into single images~\cite{wang2018dataset}. 
Specifically, these methods initialize a small amount of learnable image tensors and update them through matching the training gradients~\cite{zhao2020dc,idc}, embedding distributions~\cite{dm,wang2022cafe} or training trajectories~\cite{mtt,ftd} with the original images. 

Although the optimization-based methods achieve considerable performance as well as compression ratio, the distillation process itself still requires a large amount of time. 
We analyze the problem from the strategy of selecting original images for matching, which is mostly set as random sampling in previous works~\cite{zhao2020dc}. 
We argue that random sampling overlooks the evenness of the selected sample distribution. 
On the one hand, the matching optimization may be overly prone to certain samples with dominant matching targets, such as boundary samples with larger training gradients~\cite{wang2022cafe}. 
On the other hand, the sample diversity inside a mini-batch is also not constrained, leading to potential information insufficiency.
These factors together result in optimization instability of the dataset distillation process, and degrade the training efficiency. 

Accordingly, we propose a novel matching strategy named as \textbf{D}ataset distillation by \textbf{RE}present\textbf{A}tive \textbf{M}atching (DREAM) to address the aforementioned training efficiency issue. 
Specifically, a clustering process inside each class is conducted at intervals to generate sub-clusters reflecting the sample distribution. 
The sub-cluster centers, which not only are representative for surrounding samples, but also evenly cover the whole class distribution, are selected for matching. 
As shown in Fig.~\ref{fig:gradient}, the gradient distribution of the selected samples contains less variation. 
By only matching with representative samples, DREAM largely reduces the instability during training, and provides a smoother and more robust distillation process. 
For the synthetic image initialization, we adopt a similar clustering-based strategy, where the center sample is selected from each sub-cluster, which further accelerates the training process. 

DREAM can be easily plugged into popular dataset distillation frameworks. 
Compared with commonly adopted random matching, DREAM significantly improves the training efficiency in the distilling process.
We conduct extensive experiments to validate that
it only takes less than one eighth of the iterations for DREAM to obtain comparable performance with the baseline methods. 
In addition, given sufficient training iterations, DREAM further boosts the performance to surpass other state-of-the-art methods.

Our main contributions are summarized as:
\begin{itemize}
    \item We analyze the training efficiency of optimization-based dataset distillation from the strategy of selecting original samples for matching. 
    \item We propose a Dataset distillation by REpresentAtive Matching (DREAM) strategy. By only matching representative images, DREAM accelerates the training process by more than 8$\times$ without performance drop. 
    \item DREAM is able to be easily plugged into a variety of dataset distillation frameworks. Extensive experiments prove that DREAM consistently improves the performance of the distilled dataset.
\end{itemize}

%% file: sections/related.tex
\section{Related Works}
\subsection{Dataset Distillation}
Dataset distillation can be roughly divided into 2 categories: coreset-based and optimization-based. 

\textbf{Coreset-based methods} select a certain proportion of data based on certain metrics~\cite{guo2022deepcore,coleman2019selection}. 
Lapedriza \textit{et al}. measure the importance of the sample by the benefits obtained from training the model on the sample~\cite{lapedriza2013all}.
Toneva \textit{et al}. find that samples have different forgetting characteristics and the easily forgotten samples have larger information amount~\cite{toneva2018empirical}.
Coresets are also utilized to solve continual learning~\cite{rebuffi2017icarl,aljundi2019gradient,wiewel2021condensed} and active learning tasks~\cite{sener2017active}. 
Besides, Shleifer \textit{et al}. accelerate the search of neural network architecture by selecting a group of ``easier'' samples~\cite{shleifer2019using}.
Although coreset-based methods are practical to apply, it is hard to obtain rich information from a small amount of original samples. 
Therefore, coreset-based methods are restricted from further reducing the compression ratio. 

\textbf{Optimization-based methods} implement dataset distillation by synthesizing image samples constrained by various optimization targets. 
Wang \textit{et al}. raise the dataset distillation concept from the optimization aspect, and update the synthetic images in a meta-learning style~\cite{wang2018dataset}. 
Multiple works are then proposed to constrain the image generation by matching training gradients~\cite{zhao2020dc,zhao2021dsa,jiang2022delving}, embedding distributions~\cite{dm,wang2022cafe} and training trajectories~\cite{mtt} with original images. 
IDC injects more information into synthetic samples under the limit of fixed storage size~\cite{idc}.
Nguyen \textit{et al}. build up a distributed meta-learning framework and incorporate the kernel approximation methods~\cite{nguyen2021dataset}. 
RFAD speeds up the computation by introducing a random feature approximation~\cite{loo2022efficient}. 
HaBa employs data hallucination networks to construct base images and improves the representation capability of distilled datasets~\cite{liu2022dataset}. 
FRePo introduces an efficient meta-gradient computation method and a ``model pool'' to alleviate the overfitting~\cite{zhou2022dataset}. 
DiM~\cite{wang2023dim} transfers knowledge by distilling datasets into generative models.
Optimization-based methods largely improve the compression ratio via fusing more information into synthetic images. 
However, recent state-of-the-art methods require a large number of iterations to obtain desired validation accuracy, indicating low training efficiency. 
In this work, we focus on designing a novel matching strategy for more efficient dataset distillation training. 

\begin{figure*}[t]
\centering
\small
\begin{subfigure}{1.0\textwidth}
\centering
    \includegraphics[width=0.8\textwidth]{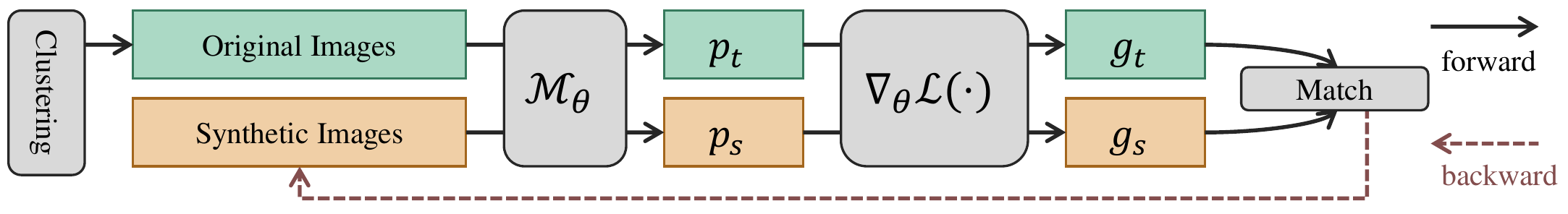}
    \caption{The training pipeline of the proposed DREAM strategy.}
    \label{fig:pipeline}
\end{subfigure}
\begin{subfigure}{0.32\textwidth}
    \includegraphics[width=\textwidth]{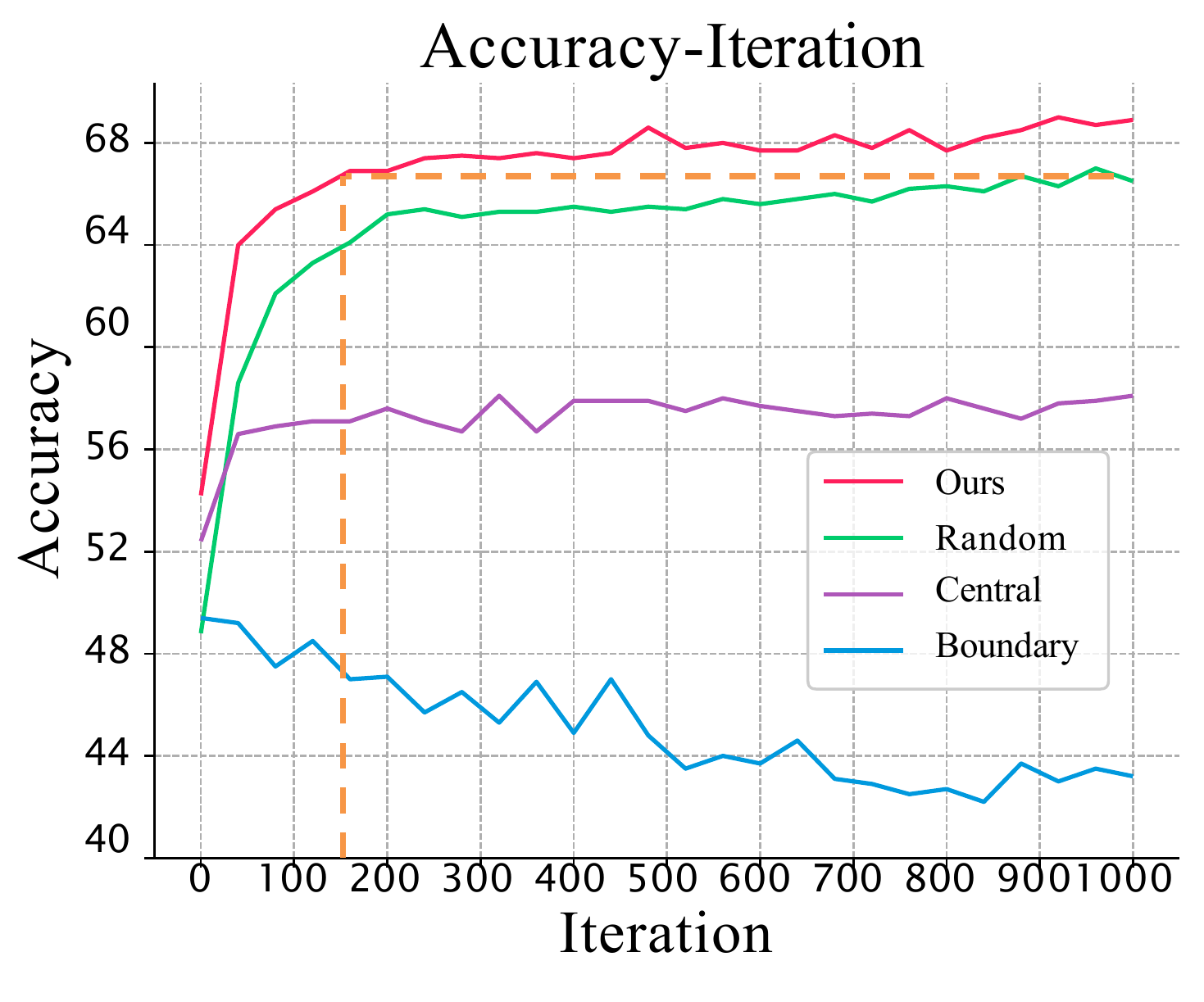}
    \caption{The accuracy curve with different strategies for selecting original images.}
    \label{fig:acc}
\end{subfigure}
\hskip 0.5em
\begin{subfigure}{0.32\textwidth}
    \includegraphics[width=\textwidth]{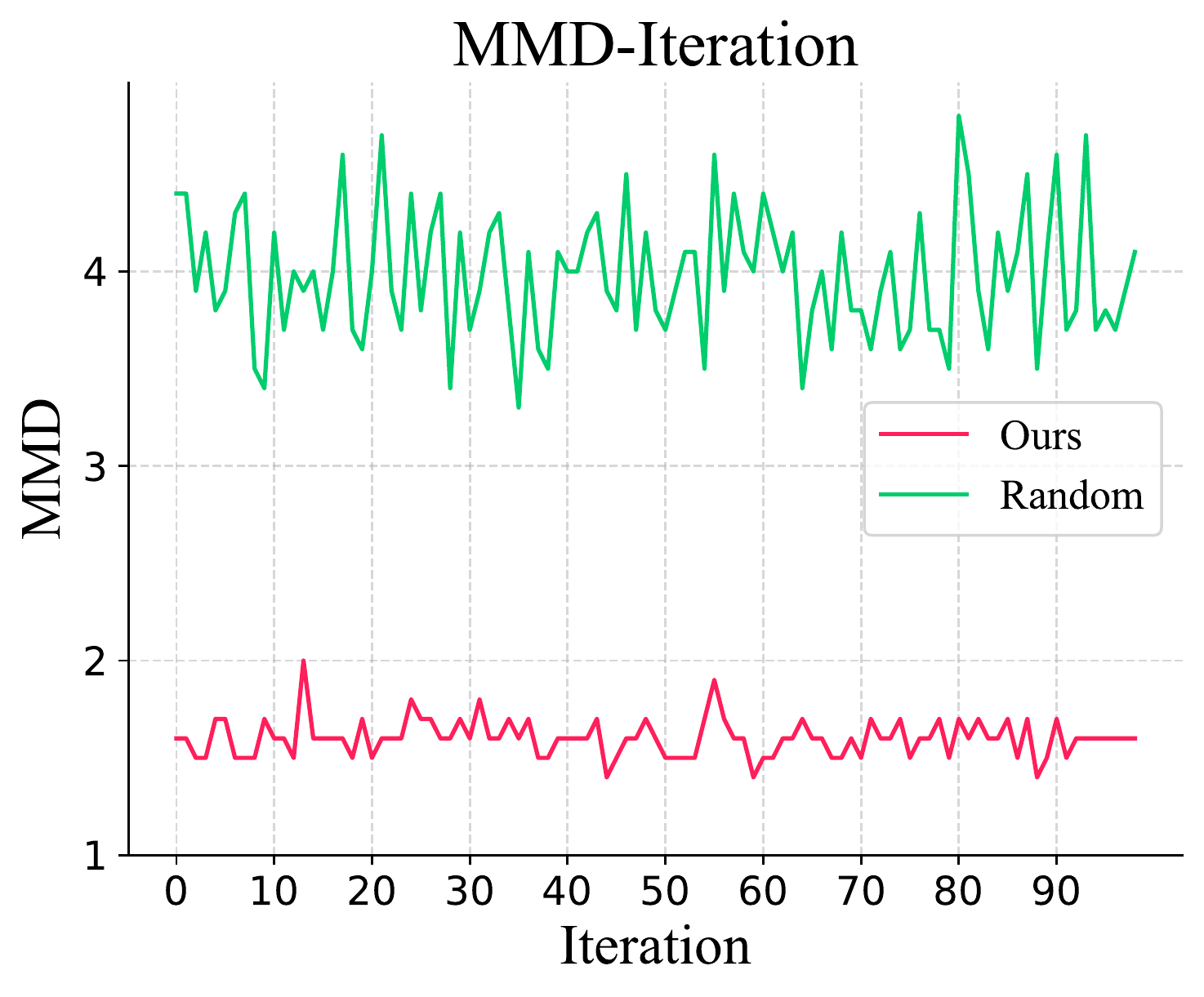}
    \caption{The MMD curve between the sampled mini-batch and the corresponding class data.}
    \label{fig:mmd}
\end{subfigure}
\hskip 0.5em
\begin{subfigure}{0.32\textwidth}
    \includegraphics[width=\textwidth]{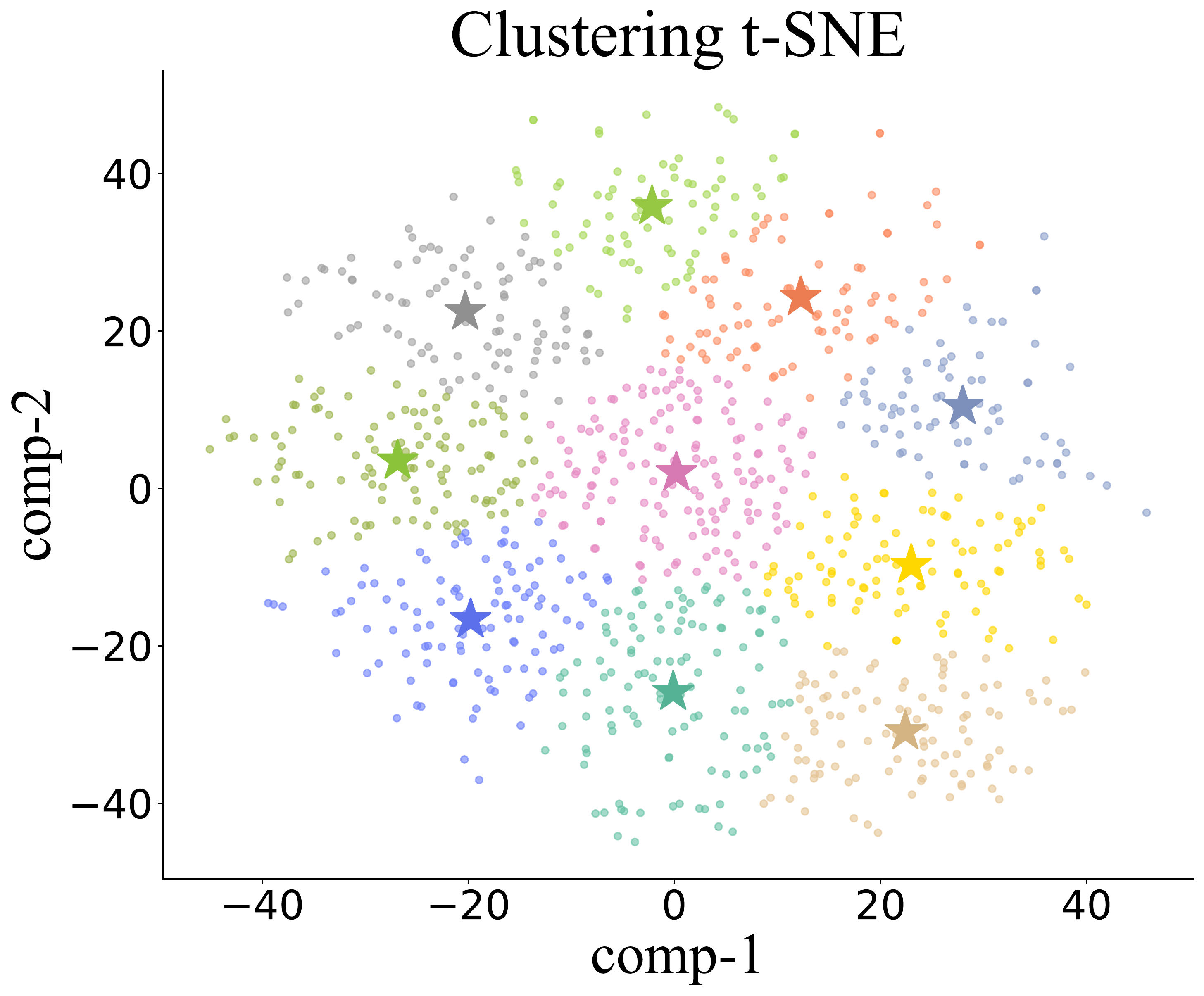}
    \caption{Example clustering and sub-cluster center results of DREAM. }
    \label{fig:cluster}
\end{subfigure}
\caption{The original images obtained by random sampling have uneven distributions, which may result in noisy or biased matching targets. Besides, the coverage of random sampling on the whole sample space is low and has large fluctuations during training. Comparatively, the centers selected by DREAM (stars) are representative for corresponding sub-clusters, and are evenly distributed over the whole class feature space. Experiments for (b) and (c) are conducted under 10 images-per-class setting on CIFAR-10. Best viewed in color. }
\label{fig:method}
\end{figure*}

\subsection{Clustering}
Clustering divides samples into groups in an unsupervised manner~\cite{rehioui2016denclue}.
K-means~\cite{k-means,k-means++} specifies the number of target clusters, and optimizes the partition to obtain clusters with similar sizes~\cite{intra-distance}.
DBSCAN, based on density, does not require the number of target clusters in advance. The clusters are formed by gradually adding data points within the tolerance range.~\cite{dbscan}.
It is applicable to dataset of any shape, yet the size of the generated clusters is unstable, outliers are excluded from clusters, and close clusters may be merged.
Hierarchical clustering methods include Agglomerative and Divisive. The former fuses multiple clusters until a certain condition is met, and the latter divides a cluster through segmentation~\cite{hierarchical}.

%% file: sections/method.tex
\section{Method}
Aiming at addressing the training efficiency problem for dataset distillation tasks, we propose a novel Dataset distillation by REpresentAtive Matching (DREAM) strategy. By only matching the representative original images, DREAM reduces the optimization instability, and achieves a smoother and more robust training process. In this section, we orderly introduce the basic training schemes of dataset distillation, our observations on the training efficiency and the detailed design of DREAM. 

\subsection{Preliminaries}
Given a large-scale dataset $\mathcal{T}=\{(\bm{x}_t^i,y_t^i)\}_{i=1}^{|\mathcal{T}|}$, the target of dataset distillation is generating a small surrogate dataset $\mathcal{S}=\{(\bm{x}_s^i,y_s^i)\}_{i=1}^{|\mathcal{S}|}$ with as little information loss as possible, where $|\mathcal{S}|\ll |\mathcal{T}|$. 
The information loss is usually measured by the performance drop between training a model with the original images $\mathcal{T}$ and the surrogate set $\mathcal{S}$. 
The commonly adopted optimization-based methods follow a synthetic pipeline. The surrogate set $\mathcal{S}$ is first initialized with random original images from $\mathcal{T}$. Under the constraints of matching objectives $\phi(\cdot)$, the synthetic images are updated to mimicking the distribution of the original images, which is formulated as:
\begin{equation}
    \mathcal{S}^*=\arg\min_\mathcal{S}\mathbf{D}\left(\phi(\mathcal{S}),\phi(\mathcal{T})\right),
\end{equation}
where $\mathbf{D}$ is the matching metric. 
Typically, we select the training gradients as the matching objective $\phi(\cdot)$. Given a random model $\mathcal{M}_\theta$ with training parameters $\theta$, $\mathcal{S}$ is supposed to give similar gradients to $\mathcal{T}$ throughout the training process of $\mathcal{M}_\theta$, that is:
\begin{equation}
    \label{eq:match}
    \mathcal{S}^*=\arg\min_\mathcal{S}\mathbf{D}\left(
    \nabla_\theta\mathcal{L}(\mathcal{M}_\theta(\mathcal{A}(\mathcal{S}))),
    \nabla_\theta\mathcal{L}(\mathcal{M}_\theta(\mathcal{A}(\mathcal{T})))\right),
\end{equation}
where $\mathcal{L}(\cdot,\cdot)$ is the training loss, and $\mathcal{A}$ is the differentiable augmentation~\cite{karras2020training,zhao2020differentiable,tran2020towards,zhao2020image}.
Practically, the matching objectives are calculated on the synthetic images and a mini-batch of original images $\{(\bm{x}_t^i,y_t^i)\}_{i=1}^{N}$ sampled from $\mathcal{T}$ with the same class labels. The objective matching and $\mathcal{M}_\theta$ training is conducted alternatively, such that gradients at different training stages are matched, which forms the inner optimization loop. The inner loop is iterated with different random $\mathcal{M}_\theta$ for more varied matching gradients. 

Recent literature offers various matching objectives and achieves significant testing accuracy via training in the small synthetic dataset~\cite{wang2022cafe,mtt,idc}. 
However, the distillation process itself still requires a large amount of training time, indicating low training efficiency. 
We analyze the relationship between the training efficiency and the sampled original images for matching, and accordingly propose a novel matching strategy. 

\subsection{Observations on Training Efficiency}

In the dataset distillation process, the knowledge is distilled from sampled original images by matching certain objectives. 
The selection of original images thereby has large influences on the training efficiency.
Recent literature usually adopts random sampling for selecting the original images~\cite{zhao2020dc,idc}. 
We set gradient matching as an example and carefully illustrate that random sampling disturbs efficient training of the dataset distillation.

Firstly, we analyze the matching effects of samples in different regions. 
Among all the samples in a class, those near the distribution center have higher prediction accuracy, indicating smaller backward gradients, while those on the decision boundaries have the contrary condition. 
For gradient matching, the central samples provide less effective supervision, while the gradients of the boundary ones largely dominate the optimization direction. 
We show the training accuracy curve of matching the synthetic images with only the central or boundary samples in Fig.~\ref{fig:acc}. 
The small gradients provided by central samples soon fail to provide effective supervision. 
On the other hand, although the boundary samples are essential for building decision boundaries, only matching with them brings chaotic matching targets, which degrades the distillation performance.

Secondly, we demonstrate that random sampling cannot guarantee an evenly distributed mini-batch along the training process. 
We record the Maximum Mean Discrepancy (MMD) between the selected mini-batch and the whole class distribution during training in Fig.~\ref{fig:mmd}. 
It can be observed that the MMD is kept at a relatively high level, with large fluctuations during the training process. 
For the gradient matching, as the mini-batch cannot effectively and consistently cover the original class sample distribution, the gradient difference of different samples are not balanced. 
The matching target of a mini-batch may be biased by boundary samples with larger training gradients, which results in unstable supervision. 

Besides, an unevenly distributed mini-batch also indicates relatively poor sample diversity. 
Information redundancy at dense regions and information lack at sparse regions make the mini-batch insufficient to represent the original data. 
The above factors result in optimization instability of the distillation process, and hence degrade the training efficiency. 
Since randomly sampling original images disturbs the training efficiency for dataset distillation training, we propose to design a novel strategy to construct mini-batches with even and diverse distribution for matching. 

\subsection{Representative Matching}

Based on the purpose of achieving stable and fast optimization, only representative original images are selected for gradient matching. 
The selection of representative images are supposed to obey the following two principles. 
On the one hand, the selected images should be evenly distributed to avoid biased matching targets. 
On the other hand, while ensuring diversity, the selected samples should reflect the overall sample distribution of the class as accurately as possible.

Therefore, we employ a clustering process for selecting representative original images. 
Out of the considerations of uniform sub-cluster sizes and distribution, without loss of generality, we adopt K-Means~\cite{k-means,k-means++,Omer_fast-pytorch-kmeans_2020} for dividing sub-clusters.
As shown in Fig.~\ref{fig:cluster}, the clustering is conducted inside each class to generate $N$ sub-clusters that reflect the sample density. $N$ is a pre-defined hyper-parameter for the mini-batch size of real images.  The sub-cluster centers evenly cover the sample space of the whole class, and simultaneously provide sufficient diversity, which perfectly meets the above principles. 

The complete training pipeline is illustrated in Fig.~\ref{fig:pipeline}.
The clustering-selected original mini-batch and the synthetic images with the same class label are passed through the random model $\mathcal{M}_\theta$ to obtain prediction scores $p_t$ and $p_s$. Subsequently calculate the classification losses and their corresponding gradients. 
The gradient differences are backwarded to update synthetic images according to Eq.~\ref{eq:match}. 
Considering the brought extra time cost, the clustering process is conducted every $I_{int}$ iterations. 

Additionally, at the beginning of the training process, we cluster the data of each class into sub-clusters corresponding to the pre-defined images-per-class number. We select the center samples of each sub-cluster as the initialization of the synthetic images. 
A more balanced clustering-based initialization better reflects the data distribution, and accelerates the convergence from the very beginning of the training process. 

%% file: sections/experiment.tex
\section{Experiments}
\input{tables/table_sota.tex}
\subsection{Datasets and Implementation Details}
We verify the effectiveness of our method on multiple popular dataset distillation benchmarks, including CIFAR10~\cite{krizhevsky2009learning}, CIFAR100~\cite{krizhevsky2009learning}, SVHN~\cite{netzer2011reading}, MNIST~\cite{lecun1998gradient}, FashionMNIST~\cite{xiao2017fashion} and TinyImageNet~\cite{deng2009imagenet}.
For evaluation, we train a model on the distilled synthetic images and test it on the original testing images. Top-1 accuracy is reported to show the performance. 

Without specific designation, the experiment is conducted on 3-layer convolutional networks (ConvNet-3)~\cite{gidaris2018dynamic} with 128 filters and instance normalization~\cite{ulyanov2016instance}. 
The matching mini-batch size for original images is set as 128. 
By default we set IDC~\cite{idc} as the baseline method. 
The gradient matching metric $\mathbf{D}$ in Eq.~\ref{eq:match} is empirically set as the mean squared error for CIFAR-10, CIFAR-100, TinyImageNet and SVHN. For MNIST and FashionMNIST, $\mathbf{D}$ is set as the mean absolute error~\cite{idc}.
We conduct 1,200 matching iterations in total, inside each of which 100 inner loops are conducted. 
SGD is set as the optimizer, with a learning rate of 0.005. 
For clustering, we employ the matching model for feature extraction. 
The clustering interval $I_{int}$ is set as 10 iterations, whose sensitiveness is analyzed in Sec.~\ref{ana:interval}. 
We also analyze the influence of different sampling strategy from the sub-clusters in Sec.~\ref{ana:sampling}. 
For evaluation, we train a network for 1,000 epochs on the distilled images with a learning rate of 0.01. 
We perform 5 experiments and report the mean and standard deviation of the results. 

\input{tables/table_tiny.tex}
\subsection{Comparison with State-of-the-art Methods}
We compare the distilled synthetic dataset performance of DREAM and other state-of-the-art (SOTA) coreset-based and optimization-based methods on multiple datasets with different images-per-class (IPC) settings in Tab.~\ref{tab:sota}.
Besides, on TinyImageNet, we compare DREAM with DM~\cite{dm} and MTT~\cite{mtt} in Tab.~\ref{tab:tiny}. 
Under all experiment circumstances, the proposed DREAM consistently surpasses other SOTA methods.
With a small IPC setting especially, under the guidance of proper gradients, DREAM is more robust than other methods, which proves the effectiveness of the representative matching strategy. 
Further narrowing the performance gap between small-scale distilled datasets and the original ones indicates that the information loss of dataset distillation is reduced. 
More detailed comparisons are included in the supplementary material. 

\subsection{Ablation Study and Analysis}
Extended experiments are designed to verify the effectiveness of our proposed DREAM strategy. 
Without specific designation, the experiment is conducted under the 10 IPC setting on CIFAR-10 dataset. 

\input{tables/table_component.tex}

\textbf{Component Combination Evaluation. }
\label{ana:component}
Firstly, we verify the isolated effects of each component in our proposed DREAM strategy in Tab.~\ref{tab:component}. 
Under the same initialization, our proposed representative matching strategy largely improves the final dataset performance. 
Comparatively, the clustering-based initialization offers a large performance lead before the training begins, yet eventually brings limited improvements. 
Nevertheless, it still provides stable boosts and accelerates the training convergence added on the representative matching to form the whole DREAM method. 
Combining all the components, the full DREAM method cuts the required iterations to achieve the baseline performance by more than 8 times. 

\input{tables/table_crossarch.tex}

\input{tables/table_sampling.tex}

Additionally, in Fig.~\ref{fig:method} we further illustrate the effectiveness of DREAM. 
Fig.~\ref{fig:acc} shows that by simply assigning samples from the sub-clusters as initialization of synthetic images, the validation performance surpasses random initialization by a large margin. 
Under the joint effect of representative matching and clustering-based initialization, DREAM achieves the final performance of random sampling with less than eighth of training iterations, demonstrating a significant training efficiency improvement. 
Continuing increasing the training iterations, DREAM further improves the dataset performance by applying proper gradient as supervision. 

From the sample distribution perspective, Fig.~\ref{fig:mmd} demonstrates that the original images selected by DREAM consistently show lower MMD scores with the original distribution with less fluctuations, compared with random sampling. 
The smaller fluctuations validates that sub-cluster centers effectively and stably cover the feature distribution, and reduces the noise at the sample level during the training process. 
With sufficient sample diversity, distribution evenness and appropriate gradient supervision, DREAM ensures a smoother and more robust optimization process for dataset distillation training. 

For better illustration of the universality of DREAM, we apply the representative matching and clustering-based initialization to some other baseline methods and receive similar effects in Tab.~\ref{tab:component}. 
The accuracy curve comparisons are presented in the supplementary material. 
It proves that DREAM is able to be easily plugged into dataset distillation frameworks and help improve the training efficiency. 

\begin{figure}[t]
    \centering
    \includegraphics[width=0.45\textwidth]{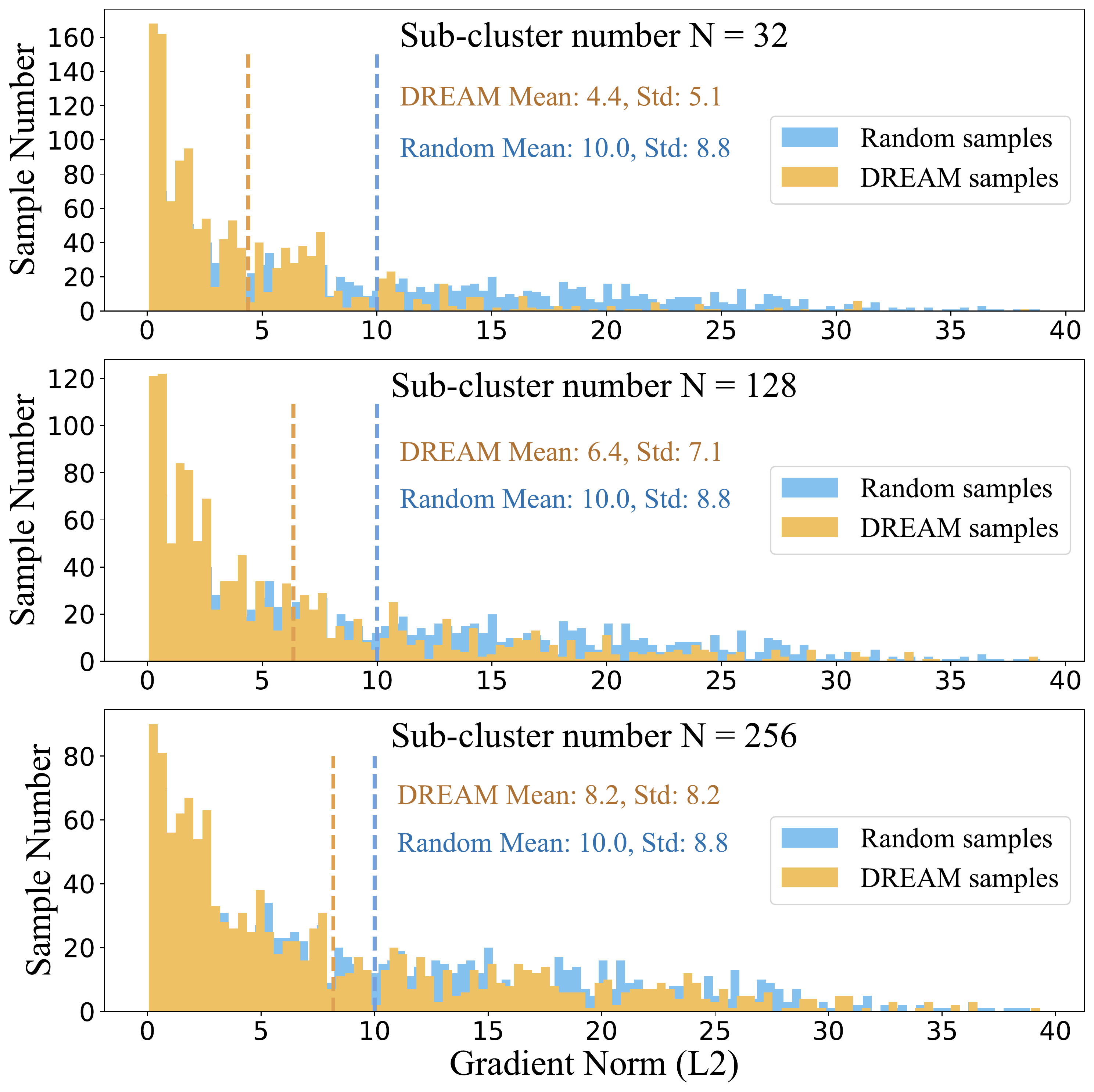}
    \caption{The gradient distribution comparison between random sampling and our proposed DREAM strategy under different sub-cluster sample number $N$. Best viewed in color.  }
    \label{fig:sampling}
\end{figure}

\begin{figure*}
\centering
\resizebox{0.19\textwidth}{0.27\textwidth}{
\begin{subfigure}{0.19\textwidth}
    \includegraphics[width=\textwidth]{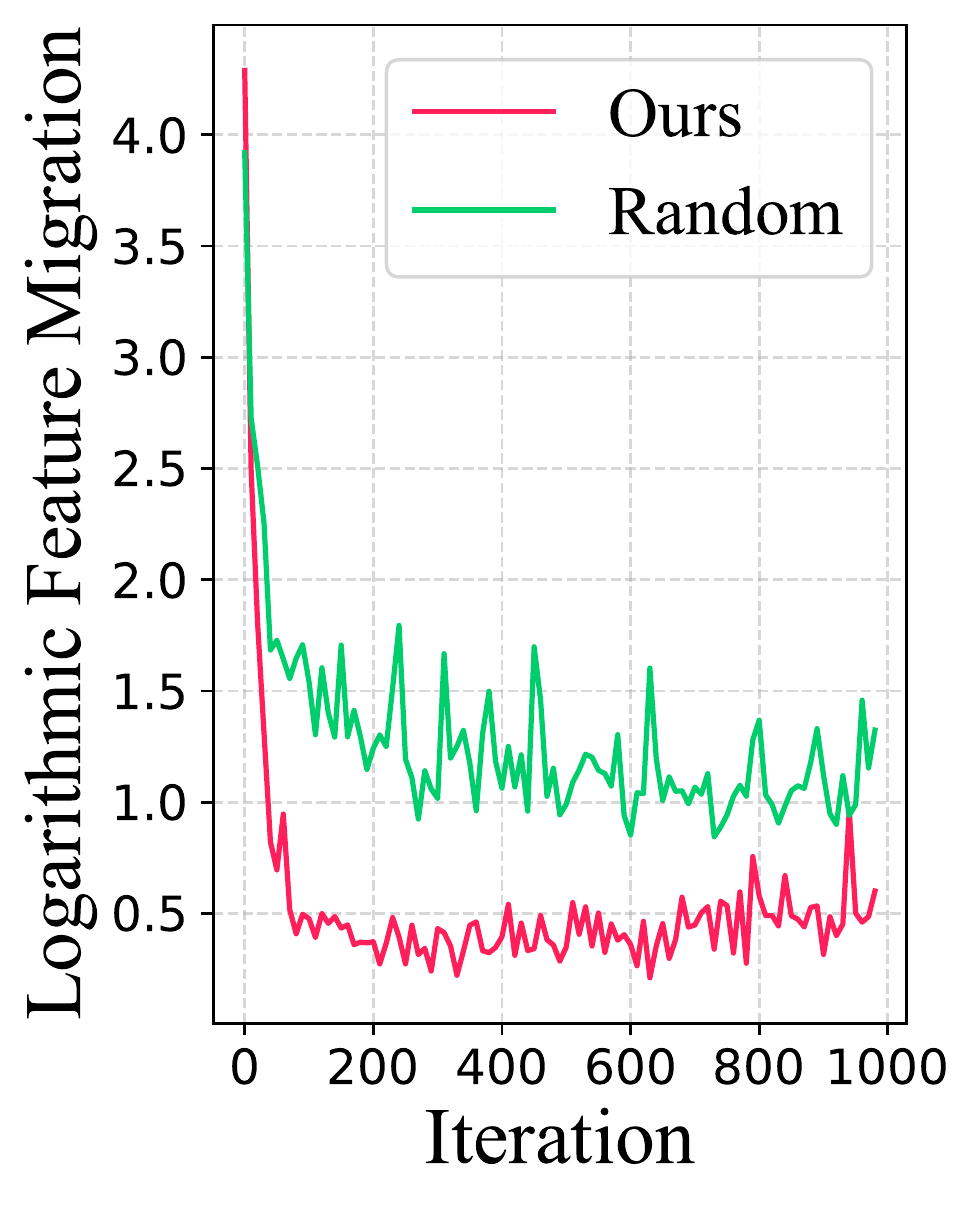}
    \caption{}
    \label{fig:osci}
\end{subfigure}
}
% \raisebox{0.08cm}{
\resizebox{0.24\textwidth}{0.273\textwidth}{
\begin{subfigure}{0.24\textwidth}
    \includegraphics[width=\textwidth]{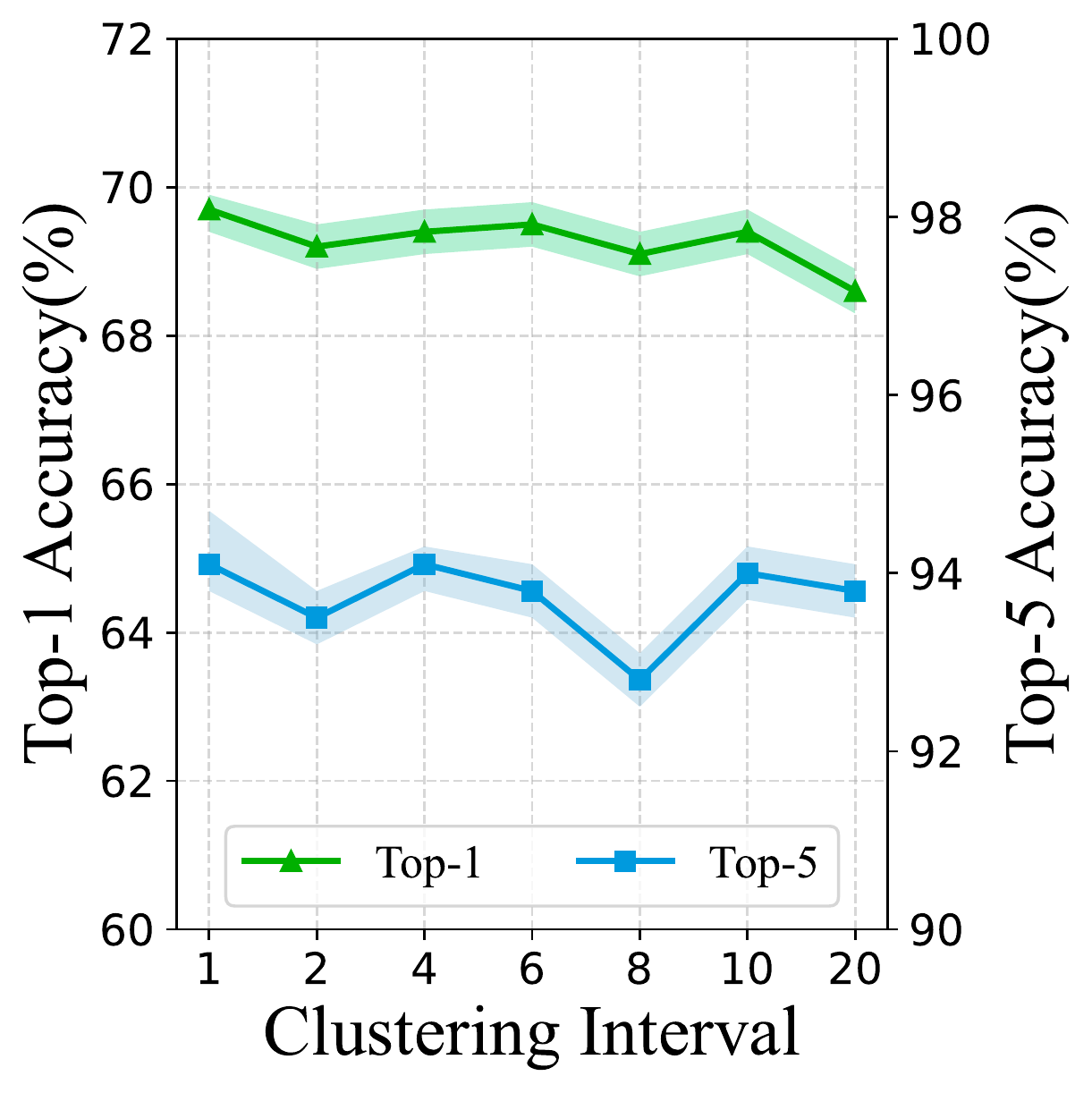}
    \caption{}
    \label{fig:interval}
\end{subfigure}
}
% }
\resizebox{0.29\textwidth}{0.27\textwidth}{
\begin{subfigure}{0.29\textwidth}
    \includegraphics[width=\textwidth]{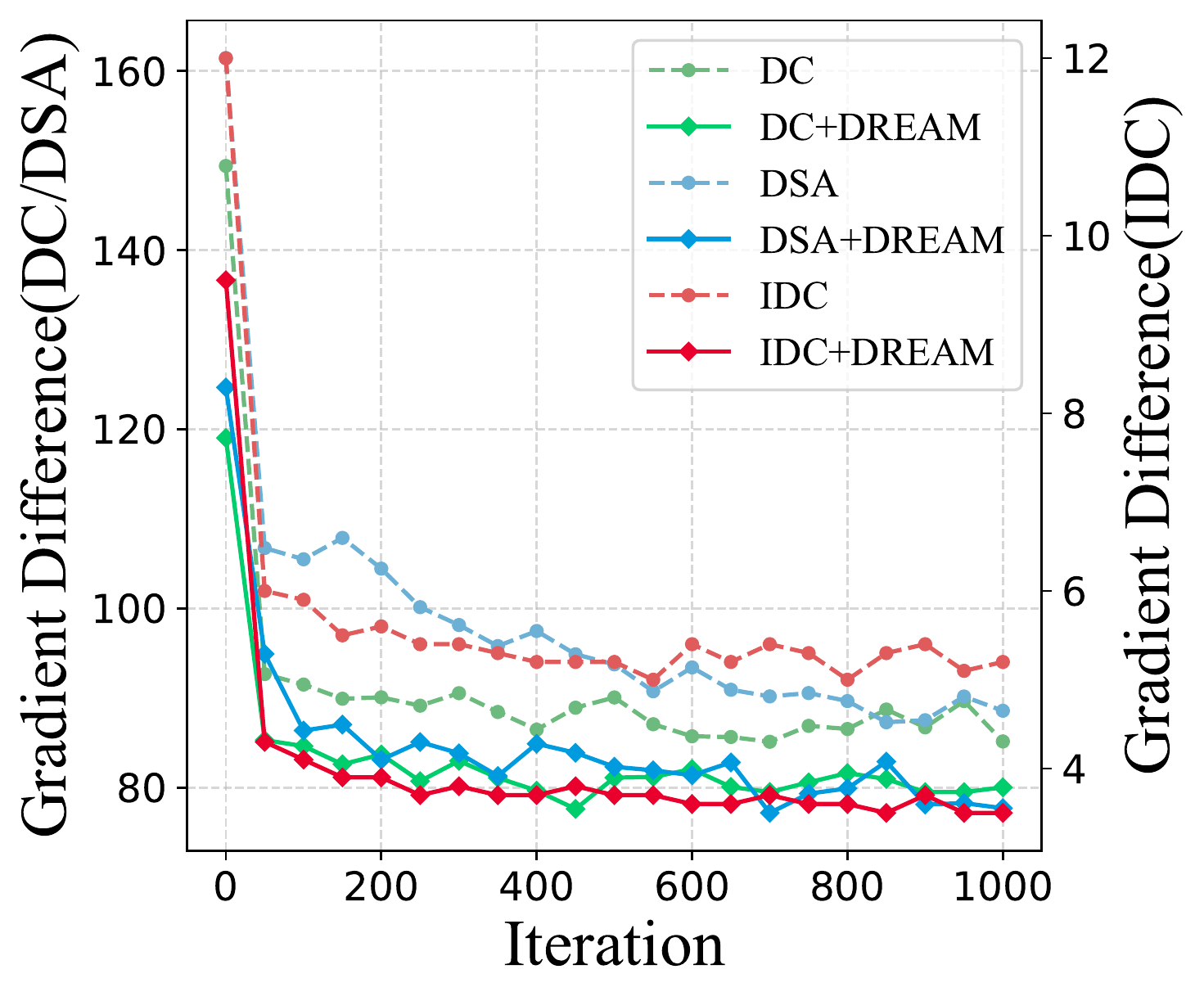}
    \caption{}
    \label{fig:loss-curve}
\end{subfigure}
}
\resizebox{0.24\textwidth}{0.273\textwidth}{
\begin{subfigure}{0.24\textwidth}
    \includegraphics[width=\textwidth]{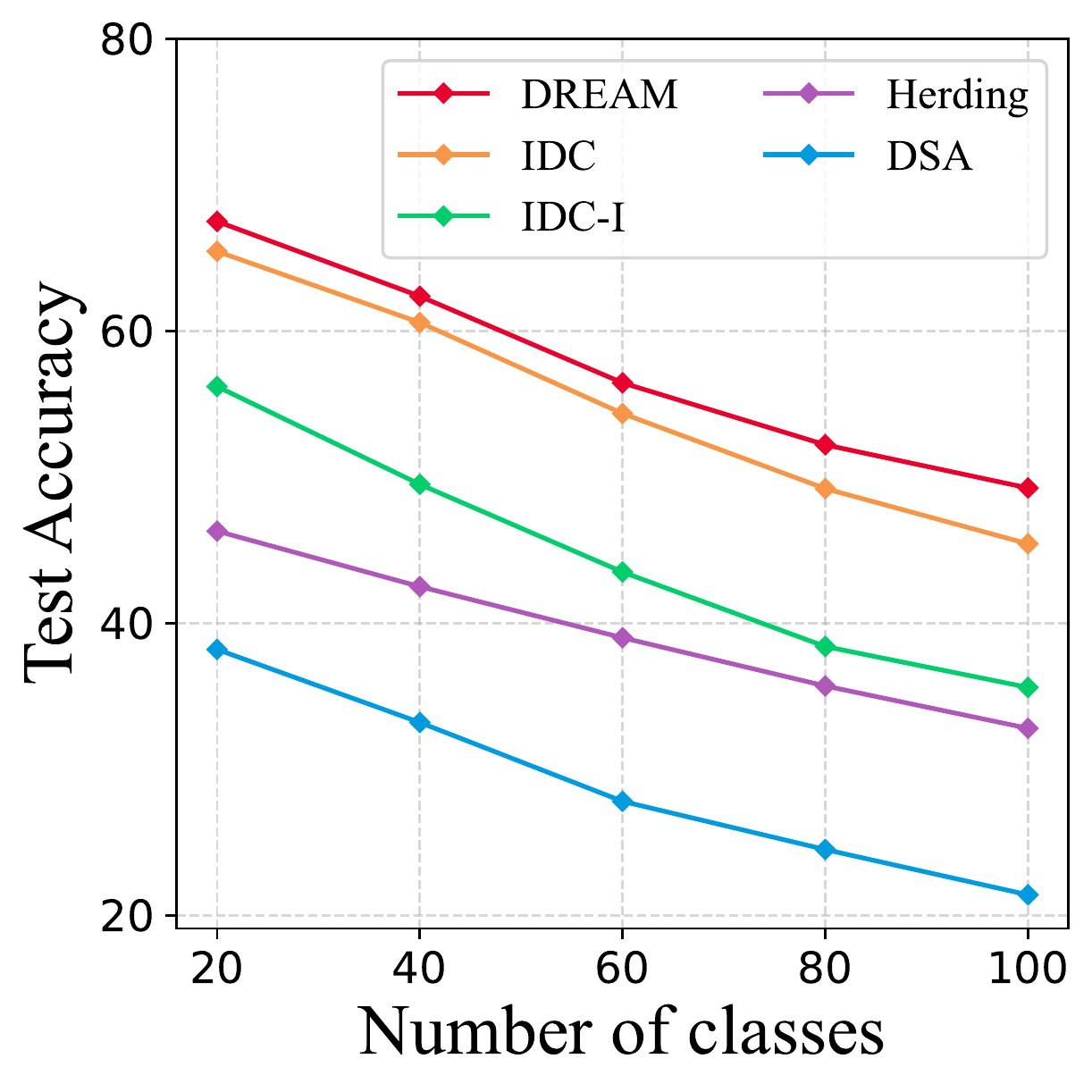}
    \caption{}
    \label{fig:continual}
\end{subfigure}
}
\label{fig:ablation}
\caption{(a): The feature migration during the training process. (b): Ablation study on different clustering interval. (c): The training loss curve during the training process. (d): The continual learning accuracy curve.  }
\end{figure*}

\textbf{Cross Architecture Generalization Analysis. }
It has been a problem for previous optimization-based dataset distillation works to generalize across architectures as the synthetic images would over-fit to the model utilized for gradient matching~\cite{zhao2020dc,idc}. 
In Tab.~\ref{tab:crossarch} we demonstrate the cross architecture performance of our proposed DREAM strategy. 
We distill the dataset with ConvNet-3 and ResNet-10~\cite{he2016deep}, and validate the performance on ConvNet-3, ResNet-10 and DenseNet-121~\cite{huang2017densely}. 

DREAM surpasses the compared methods on both the absolute performance and the performance drop when applying the distilled dataset on an unseen architecture. 
The strong cross architecture generalization capability verifies that DREAM helps build a more reasonable distilled dataset compared to random sampling. 

\textbf{Sampling Strategy Analysis. }
\label{ana:sampling}
Representative matching conducts clustering for each class and samples original images from the sub-clusters to form a mini-batch. 
We analyze the influence of different sampling strategy on the training results in Tab.~\ref{tab:sampling} and Fig.~\ref{fig:sampling}. 
Among each sub-cluster, the top-$n$ samples closest to the center are selected. 
By grouping different sub-cluster number and samples per sub-cluster, we are able to obtain original image mini-batches different in scale and diversity. 
As observed in the results, by representative matching the dataset performance is generally stable, and receives improvements to certain extent over the baseline (67.5). 

Compared in more detail, with a small sub-cluster number $N=32$, the sub-cluster centers are more likely to be distributed in areas with smaller gradients, as shown in the first row of Fig.~\ref{fig:sampling}. 
As the random model $\mathcal{M}_\theta$ is trained, these samples gradually fail to provide effective gradients for supervision, resulting in a sub-optimal performance. 
Oppositely, a larger sub-cluster number $N=256$ involves a distribution closer to random sampling, which brings a small performance drop, as shown in the last row of Fig.~\ref{fig:sampling}. 
Due to memory limitations, it is not applicable to further increase $N$, but it is conceivable that the extreme condition should yield similar results to random sampling. 
On the other hand, the sample number per sub-cluster $n$ has only a slight effect on the results. 
The group of 1 center sample per sub-cluster and 128 sub-clusters in total is proved to obtain the optimal gradient supervision as in the second row of Fig.~\ref{fig:sampling}, and is chosen for mini-batch composition. 

\textbf{Training Stability Analysis. }
In order to more intuitively demonstrate the effects of the proposed DREAM strategy on the training process, we visualize the feature migration of DREAM and random sampling in Fig.~\ref{fig:osci}. 
Specifically, we randomly select a synthetic image as initialization and record its updated version every 10 iterations. 
We employ a random network to extract the features of all images, and calculate the Euclidean distance between adjacent versions of images. 
For DREAM, at the beginning of the training process, under the direction of proper gradients, the synthetic image goes through a larger migration.
Within 100 iterations, the synthetic image has reached a relatively optimal position, and makes subsequent fine-tuning. 
On the contrary, there are still large fluctuations for the synthetic image matched with randomly sampled original images in the late training period, partly due to the noisy matching targets generated by uneven mini-batches. 

\textbf{Clustering Interval Sensitivity Analysis. }
\label{ana:interval}
We evaluate the influence of different clustering interval $I_{int}$ on the final dataset performance in Fig.~\ref{fig:interval}. 
Conducting clustering at every iteration leads to the best performance, while adding the clustering interval until 10 brings mild influences. 
As there is an obvious top-1 accuracy degradation when the interval is further increased to 20, we select an interval of 10 to balance the distilled dataset performance and the extra calculation cost. 
More analysis on the computational cost of clustering is included in the supplementary material. 

\textbf{Experimental results on ImageNet-1K.}
We compare DREAM with the current state-of-the-art method TESLA~\cite{cui2023scaling} on ImageNet-1K in Tab.~\ref{tab:imagenet}. The experimental setting is the same as in TESLA.
DREAM shows excellent performance on large datasets.
\begin{table}[h]
\centering
\caption{Results on ImageNet.}
\vspace{-8pt}
\label{tab:imagenet}
\scriptsize
\setlength{\tabcolsep}{4pt}
\begin{tabular}{c|cc}
    \toprule
    Method & TESLA & DREAM \\
    \midrule
    Acc (IPC=10) & 17.8 & \textbf{18.5} \\
    \bottomrule
\end{tabular}
\end{table}

\textbf{Differences from DC-BENCH\cite{cui2022dc}.}
\label{ana:difference}
We provide a detailed comparison between DREAM and DC-BENCH to clarify their distinctions. DC-BENCH solely concentrates on a better initialization and lacks specific designs for the subsequent matching-based optimization, while DREAM selects representative samples for matching and enables the realization of a fully efficient training process for distillation. Furthermore, DREAM conducts extensive experiments to analyze the impact of cluster number, sample number per cluster and clustering interval. DC-BENCH achieves comparable performance using 30$\%$ of iterations, whereas DREAM achieves similar results with only 10-20$\%$ iterations. Additionally, given sufficient training time, DREAM further achieves up to 3.7$\%$ and 5.8$\%$ accuracy improvements for gradient matching and distribution matching respectively which surpass DC-BENCH's 1.3$\%$.
\subsection{Visualizations}

\textbf{Gradient Difference Curve.}
As the dataset distillation training is constrained by matching the training gradients, a smaller gradient difference also indicates a better matching effect. 
Therefore, we also visualize the gradient difference curve of the dataset distillation in Fig.~\ref{fig:loss-curve}, which is calculated by the training loss Eq.~\ref{eq:match}. 
We add the DREAM strategy to DC, DSA and IDC methods. 
Throughout the training process, DREAM holds a smaller gradient difference compared with the baseline methods. 
On the one hand, it verifies the effectiveness of DREAM on improving the training efficiency to reduce the gradient difference in limited iterations. 
On the other hand, the large fluctuations of the baseline methods also validate the existence of noisy gradients generated by random sampling. 

\begin{figure}
    \centering
    \begin{overpic}[width=0.48\textwidth]{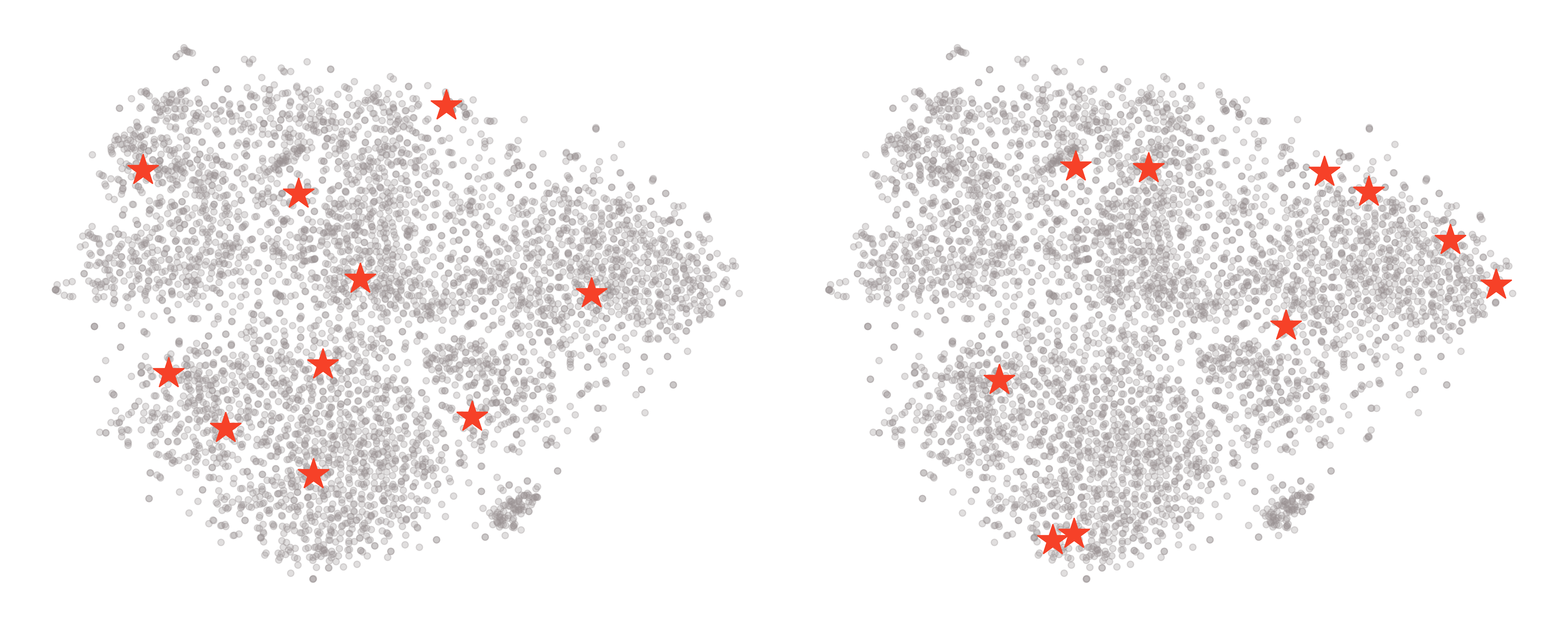}
        \put(35, 5){Ours}
        \put(85, 5){Random}
    \end{overpic}
    \caption{The sample distribution comparison on the final distilled images (marked as red stars) between our proposed DREAM (left) and random sampling (right). }
    \label{fig:distribution}
\end{figure}

\begin{figure}
\centering
\includegraphics[width=\columnwidth]{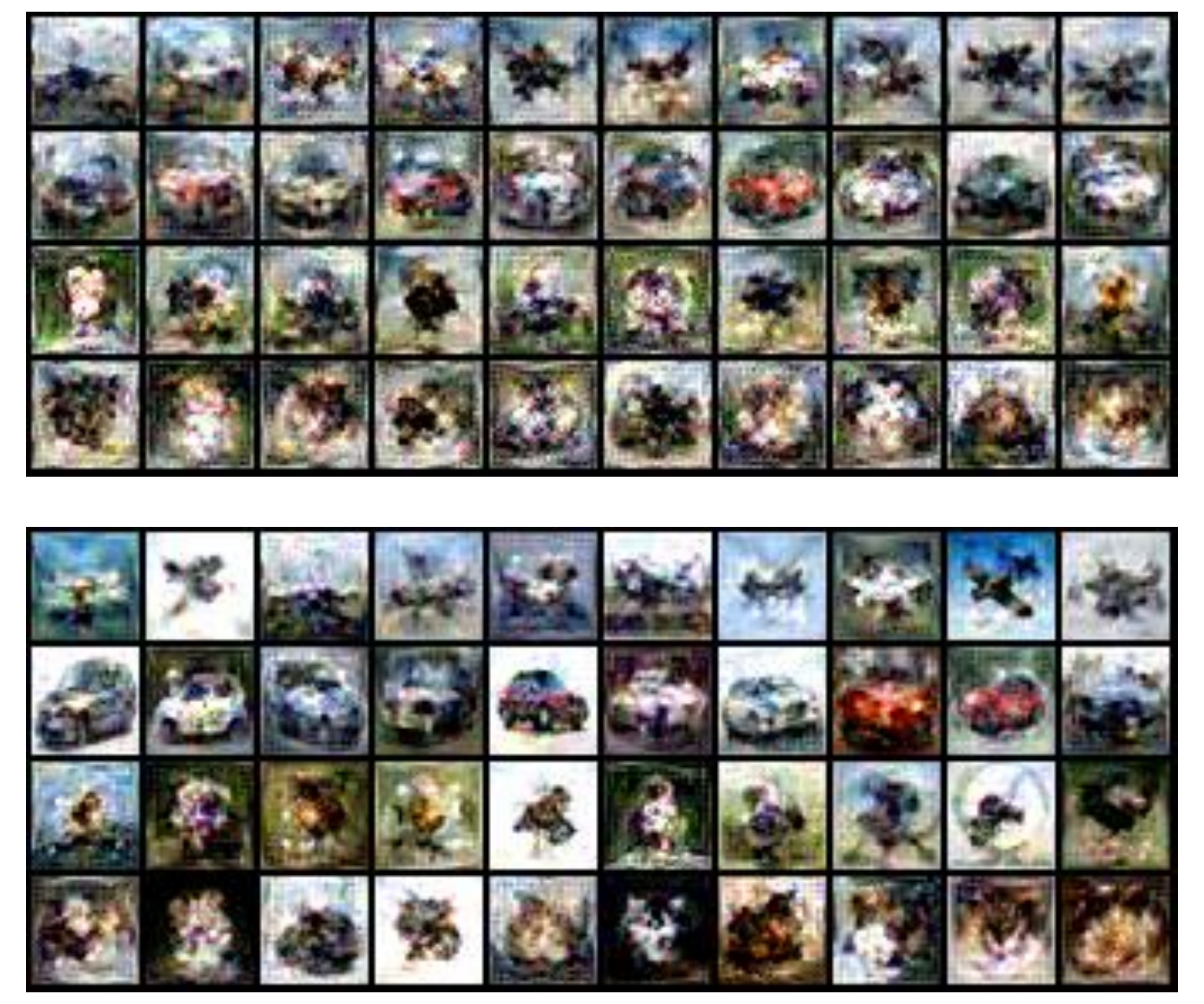}
\caption{The distilled dataset comparison between DC (Upper row) and DC with DREAM strategy (Bottom row) on CIFAR-10 (plane, car, dog, cat classes). DREAM introduces more obvious categorical characteristics and variety to the distilled image. Best viewed in color. More visualization is provided in supplementary material. }
\label{fig:dc-comp}
\end{figure}

\textbf{Sample Distribution Visualization.}
In order to more intuitively demonstrate the effectiveness of our proposed DREAM on generating synthetic sets well covering the original sample distribution, we visualize the t-SNE graphs of the synthetic images for both random sampling and DREAM. 
As shown in Fig.~\ref{fig:distribution}, the final distribution constrained by DREAM strategy evenly cover the whole class, while random sampling generates biased optimization results. 
Furthermore, a large percentage of samples are pulled to the distribution edge in the random sampling results, which also validates that the matching is biased by boundary samples with larger gradients. 
By consistently providing proper gradient supervision, DREAM achieves a more diverse and robust distillation result. 

\textbf{Synthetic Image Visualization.}
In order to more intuitively demonstrate the effects of DREAM on the distilled images, we compare the distillation results of adding the proposed DREAM strategy or not in Fig.~\ref{fig:dc-comp}.
DREAM improves the quality of the distilled datasets from two perspectives. Firstly, the images optimized by DREAM show more obvious categorical characteristics. Secondly, DREAM introduces more variety to the distilled images. With these two improvements, DREAM helps the distilled datasets to obtain better validation performance. 

\subsection{Application on Continual Learning}
Dataset distillation generates compact datasets that are able to represent the original ones, which can thus be applied to continual learning problems~\cite{rebuffi2017icarl,aljundi2019gradient,wiewel2021condensed,idc}. 
We further validate the effectiveness of the proposed DREAM strategy on the continual learning scenarios in Fig.~\ref{fig:continual}.
Following the settings in~\cite{zhao2020dc,idc}, we conduct a 5-step class-incremental experiment on CIFAR-100, each step with 20 classes. 
For better demonstrating the generalization capability of DREAM, the distillation synthesis is conducted on ConvNet-3, and the validation on ResNet-10. 

DREAM consistently maintains performance advantages over other approaches throughout the training process, and the performance gap is further enlarged as the learnt class number is gradually increased. 
It proves that better distillation quality helps the model construct clearer decision boundaries and memorize the discriminative information.

%% file: tables/table_sota.tex
\begin{table*}[t]
% \vspace{-10pt}
% \renewcommand\arraystretch{0.9}
\caption{Top-1 accuracy of test models trained on distilled synthetic images on multiple datasets. The distillation training is conducted with ConvNet-3. $^\dagger$ denotes the reported error range is reproduced by us. }
\label{tab:sota}
\centering
\footnotesize
\setlength{\tabcolsep}{3.8pt}
\begin{tabular}{ccc|cc|ccccccc|c}
\toprule
\multirow{2}{*}{}& \multirow{2}{*}{IPC} & \multirow{2}{*}{Ratio \%} & \multicolumn{2}{c|}{Coreset Selection}   & \multicolumn{7}{c|}{Training Set Synthesis} & Whole \\ %\cline{4-7}
& & & Random & Herding & DC~\cite{zhao2020dc} & DSA~\cite{zhao2021dsa} & DM~\cite{dm} & CAFE~\cite{wang2022cafe} & MTT~\cite{mtt} & IDC~\cite{idc} & \textbf{DREAM} & Dataset \\ \midrule
\multirow{3}{*}{MNIST}
& 1 & 0.017 & 64.9$_{\pm3.5}$ & 89.2$_{\pm1.6}$ & 91.7$_{\pm0.5}$ & 88.7$_{\pm0.6}$ & 89.7$_{\pm0.6}$ & 93.1$_{\pm0.3}$ &-& 94.2$_{\pm0.2}$$^\dagger$ & \bf{95.7}$_{\pm0.3}$ & \\
& 10 & 0.17 & 95.1$_{\pm0.9}$ & 93.7$_{\pm0.3}$ & 97.4$_{\pm0.2}$ & 97.8$_{\pm0.1}$ & 97.5$_{\pm0.1}$ & 97.2$_{\pm0.3}$ &-& 98.4$_{\pm0.1}$$^\dagger$ & \bf{98.6}$_{\pm0.1}$ & 99.6$_{\pm0.0}$ \\
& 50 & 0.83 & 97.9$_{\pm0.2}$ & 94.8$_{\pm0.2}$ & 98.8$_{\pm0.2}$ & 99.2$_{\pm0.1}$ & 98.6$_{\pm0.1}$ & 98.6$_{\pm0.2}$ &-& 99.1$_{\pm0.1}$$^\dagger$ & \bf{99.2}$_{\pm0.1}$ &  \\ \midrule 

\multirow{3}{*}{FashionMNIST}
& 1 & 0.017 & 51.4$_{\pm3.8}$ & 67.0$_{\pm1.9}$ & 70.5$_{\pm0.6}$ & 70.6$_{\pm0.6}$ & - & 77.1$_{\pm0.9}$ &-& 81.0$_{\pm0.2}$$^\dagger$ & \bf{81.3}$_{\pm0.2}$ & \\
& 10 & 0.17 & 73.8$_{\pm0.7}$ & 71.1$_{\pm0.7}$ & 82.3$_{\pm0.4}$ & 84.6$_{\pm0.3}$ & - & 83.0$_{\pm0.4}$ &-& 86.0$_{\pm0.3}$$^\dagger$ & \bf{86.4}$_{\pm0.3}$ & 93.5$_{\pm0.1}$ \\
& 50 & 0.83 & 82.5$_{\pm0.7}$ & 71.9$_{\pm0.8}$ & 83.6$_{\pm0.4}$ & \textbf{88.7$_{\pm0.2}$} & - & 84.8$_{\pm0.4}$ &-& 86.2$_{\pm0.2}$$^\dagger$& 86.8$_{\pm0.3}$&  \\ \midrule

\multirow{3}{*}{SVHN}
& 1 & 0.014 & 14.6$_{\pm1.6}$ & 20.9$_{\pm1.3}$ & 31.2$_{\pm1.4}$ & 27.5$_{\pm1.4}$ & - & 42.6$_{\pm3.3}$ &-& 68.5$_{\pm0.9}$$^\dagger$ & \bf{69.8}$_{\pm0.8}$ & \\
& 10 & 0.14 & 35.1$_{\pm4.1}$ & 50.5$_{\pm3.3}$ & 76.1$_{\pm0.6}$ & 79.2$_{\pm0.5}$ & - & 75.9$_{\pm0.6}$ &-& 87.5$_{\pm0.3}$$^\dagger$ & \bf{87.9}$_{\pm0.4}$ & 95.4$_{\pm0.1}$ \\
& 50 & 0.7 & 70.9$_{\pm0.9}$ & 72.6$_{\pm0.8}$ & 82.3$_{\pm0.3}$ & 84.4$_{\pm0.4}$ & - & 81.3$_{\pm0.3}$ &-& 90.1$_{\pm0.1}$$^\dagger$& \bf{90.5}$_{\pm0.1}$ &  \\ \midrule

\multirow{3}{*}{CIFAR10}
& 1 & 0.02 & 14.4$_{\pm2.0}$ & 21.5$_{\pm1.2}$ & 28.3$_{\pm0.5}$ & 28.8$_{\pm0.7}$ & 26.0$_{\pm0.8}$ & 30.3$_{\pm1.1}$ & 46.3$_{\pm0.8}$ & 50.6$_{\pm0.4}$$^\dagger$ & \bf{51.1}$_{\pm0.3}$ & \\
& 10 & 0.2 & 26.0$_{\pm1.2}$ & 31.6$_{\pm0.7}$ & 44.9$_{\pm0.5}$ & 52.1$_{\pm0.5}$ & 48.9$_{\pm0.6}$ & 46.3$_{\pm0.6}$ & 65.3$_{\pm0.7}$ & 67.5$_{\pm0.5}$ & \bf{69.4}$_{\pm0.4}$ & 84.8$_{\pm0.1}$\\
& 50 & 1.0 & 43.4$_{\pm1.0}$ & 40.4$_{\pm0.6}$ & 53.9$_{\pm0.5}$ & 60.6$_{\pm0.5}$ & 63.0$_{\pm0.4}$ & 55.5$_{\pm0.6}$ & 71.6$_{\pm0.2}$ & 74.5$_{\pm0.1}$ & \bf{74.8}$_{\pm0.1}$ &  \\ \midrule

\multirow{3}{*}{CIFAR100}
& 1 & 0.2 & 4.2$_{\pm0.3}$ & 8.4$_{\pm0.3}$ & 12.8$_{\pm0.3}$ & 13.9$_{\pm0.3}$ & 11.4$_{\pm0.3}$ & 12.9$_{\pm0.3}$ & 24.3$_{\pm0.3}$ & - & \bf{29.5}$_{\pm0.3}$ & \\
& 10 & 2 & 14.6$_{\pm0.5}$ & 17.3$_{\pm0.3}$ & 25.2$_{\pm0.3}$ & 32.3$_{\pm0.3}$ & 29.7$_{\pm0.3}$ & 27.8$_{\pm0.3}$ & 40.1$_{\pm0.4}$ & 45.1$_{\pm0.4}$$^\dagger$ & \bf{46.8}$_{\pm0.7}$ & 56.2$_{\pm0.3}$ \\
& 50 & 10 & 30.0$_{\pm0.4}$ & 33.7$_{\pm0.5}$ & - & 42.8$_{\pm0.4}$ & 43.6$_{\pm0.4}$ & 37.9$_{\pm0.3}$ & 47.7$_{\pm0.2}$ & - & \bf{52.6}$_{\pm0.4}$ &  \\ \bottomrule

\end{tabular}
% \vspace{-5pt}
% \vspace{-15pt}
\end{table*}

%% file: tables/table_tiny.tex
\begin{table}[t]
\caption{Top-1 accuracy of test models trained on distilled synthetic images on TinyImageNet. The distillation training is conducted with ConvNet-3. }
\label{tab:tiny}
\centering
\small
\setlength{\tabcolsep}{4pt}
\begin{tabular}{cc|ccc|c}
\toprule
IPC & Ratio \% & DM~\cite{dm} & MTT~\cite{mtt} & \textbf{DREAM}& Whole \\ \midrule
1 & 0.017 & 3.9$_{\pm0.2}$ & 8.8$_{\pm0.3}$ & \bf{10.0}$_{\pm0.4}$ & \multirow{2}{*}{37.6$_{\pm0.4}$}\\
50 & 0.83 & 24.1$_{\pm0.3}$ & 28.0$_{\pm0.3}$ & \bf{29.5}$_{\pm0.3}$ &  \\ \bottomrule
\end{tabular}
\end{table}

%% file: tables/table_component.tex
\begin{table}[t]
\setlength{\tabcolsep}{3pt}
    \caption{Ablation study on the components of the proposed DREAM. RM indicates Representative Matching, and Init stands for clustering-based initialization. “Iter” stands for the required iterations to achieve the baseline performance.}
    \label{tab:component}
    \centering
    \small
    \begin{tabular}{lcccc|lccc}
    \toprule
        \multirow{2}{*}{}& \multicolumn{2}{c}{Comp} & \multirow{2}{*}{Top-1} & \multirow{2}{*}{Iter} & \multirow{2}{*}{}& \multicolumn{2}{c}{Comp} & \multirow{2}{*}{Top-1} \\
         & RM & Init & & & & RM & Init\\
        \midrule
         \multirow{4}{*}{IDC}& - & - & 67.5$_{\pm0.5}$ & 1000 & \multirow{2}{*}{DC} & - & - & 44.9$_{\pm0.5}$\\
          & \checkmark & - & 68.9$_{\pm0.5}$ & 350 & & \checkmark & \checkmark & \textbf{45.9}$_{\pm0.3}$\\\cline{6-9}
          & - & \checkmark & 68.1$_{\pm0.3}$ & 750 & \multirow{2}{*}{DSA} & - & - & 52.1$_{\pm0.5}$ \\
          & \checkmark & \checkmark & \textbf{69.4}$_{\pm0.4}$ & \textbf{150} & & \checkmark & \checkmark & \textbf{53.1}$_{\pm0.4}$\\
        \bottomrule
    \end{tabular}
\end{table}

%% file: tables/table_crossarch.tex
\begin{table}[t]
    \caption{Ablation study on cross architecture distilled dataset performance of the proposed DREAM strategy. The dataset is first distilled on a model D and then validated on another model T. † denotes the result is reproduced by us.}
    \label{tab:crossarch}
    \centering
    \small
    \begin{tabular}{lcccc}
    \toprule
        & $\mathrm{D}\backslash\mathrm{T}$ & Conv-3 & Res-10 & Dense-121 \\
        \midrule
        \multirow{2}{*}{MTT~\cite{mtt}} & Conv-3 & 64.3$_{\pm0.7}$ & 34.5$_{\pm0.6}$$^\dagger$ & 41.5$_{\pm0.5}$$^\dagger$\\
         & Res-10 & 44.2$_{\pm0.3}$$^\dagger$& 20.4$_{\pm0.9}$$^\dagger$ & 24.2$_{\pm1.3}$$^\dagger$\\
        \midrule
        \multirow{2}{*}{IDC~\cite{idc}} & Conv-3 & 67.5$_{\pm0.5}$ & 63.5$_{\pm0.1}$ & 61.6$_{\pm0.6}$ \\
         & Res-10 & 53.6$_{\pm0.6}$$^\dagger$ & 50.6$_{\pm0.9}$$^\dagger$ & 51.7$_{\pm0.6}$$^\dagger$\\
        \midrule
        \multirow{2}{*}{DREAM} & Conv-3 & \textbf{69.4}$_{\pm0.4}$ & \textbf{66.3}$_{\pm0.8}$ & \textbf{65.9}$_{\pm0.5}$\\
         & Res-10 & \textbf{53.7}$_{\pm0.6}$ & \textbf{51.0}$_{\pm0.9}$ & \textbf{52.8}$_{\pm0.6}$\\
        \bottomrule
    \end{tabular}
\end{table}

%% file: tables/table_sampling.tex
\begin{table}[t]
    \caption{Ablation study on different sampling strategy to form a mini-batch from sub-clusters. }
    \centering
    \small
    \setlength{\tabcolsep}{4pt}
    \begin{tabular}{lc|cccc}
    \toprule
        & & \multicolumn{4}{c}{Sub-cluster number $N$} \\
        & & 32 & 64 & 128 & 256 \\
        \midrule
         & 1 & 67.2$_{\pm0.3}$ & 68.5$_{\pm0.1}$ & \textbf{69.4}$_{\pm0.4}$ & 68.9$_{\pm0.2}$ \\
        Samples per& 2 & 67.7$_{\pm0.3}$ & 68.6$_{\pm0.3}$ & 69.2$_{\pm0.7}$ & - \\
        sub-cluster $n$ & 4 & 67.7$_{\pm0.4}$ & 68.7$_{\pm0.4}$ & - & - \\
        & 8 & 67.5$_{\pm0.3}$ & - & - & - \\
    \bottomrule
    \end{tabular}
    \label{tab:sampling}
\end{table}

%% file: sections/supp.tex
\clearpage

\section{Comparisons with More Methods}
We compare the distilled dataset performance between DREAM and more methods in Tab.~\ref{tab:other_methods}. 
The experiments are conducted under 1, 10, and 50 images-per-class (IPC) settings on MNIST~\cite{lecun1998gradient}, SVHN~\cite{netzer2011reading}, CIFAR-10, and CIFAR-100~\cite{krizhevsky2009learning} datasets.
DREAM achieves SOTA results on most cases. 
Especially when IPC is small, DREAM has gained significant performance gap over other methods, which also validates the effectiveness of matching representative samples.
RFAD~\cite{loo2022efficient} employs ConvNet with 1024 convolutional channels, while our results are reported based on 128-channel ConvNet. Except for IPC=1 CIFAR-10, DREAM distills better synthetic images than RFAD. 
HaBa~\cite{liu2022dataset} involves a data hallucination process, which generates more samples from base images. It holds higher performance on IPC=10 CIFAR-10 and IPC=1 CIFAR-100, while in other circumstances, DREAM has superior performances. 

\section{Accuracy Curve Visualization}
We apply the DREAM strategy to more dataset distillation methods, such as DC~\cite{zhao2020dc}, DSA~\cite{zhao2021dsa}, etc.
In addition to stable performance improvements, we visualize the accuracy curve during training in Fig.~\ref{fig:curve}. 
It can be observed that compared with the original methods, DREAM only requires one fifth and one tenth of the iteration number on the DC and DSA to achieve the original performance, respectively. 
Further increasing the training time brings continuous performance improvement, which also proves that our method is universal and is able to be easily plugged for popular dataset distillation frameworks.
The above experiments are all based on the setting of 10 images-per-class on CIFAR10.

\section{DREAM on Distribution Matching}
\begin{figure*}[t]
\centering
\begin{subfigure}{0.32\textwidth}
    \includegraphics[width=\textwidth]{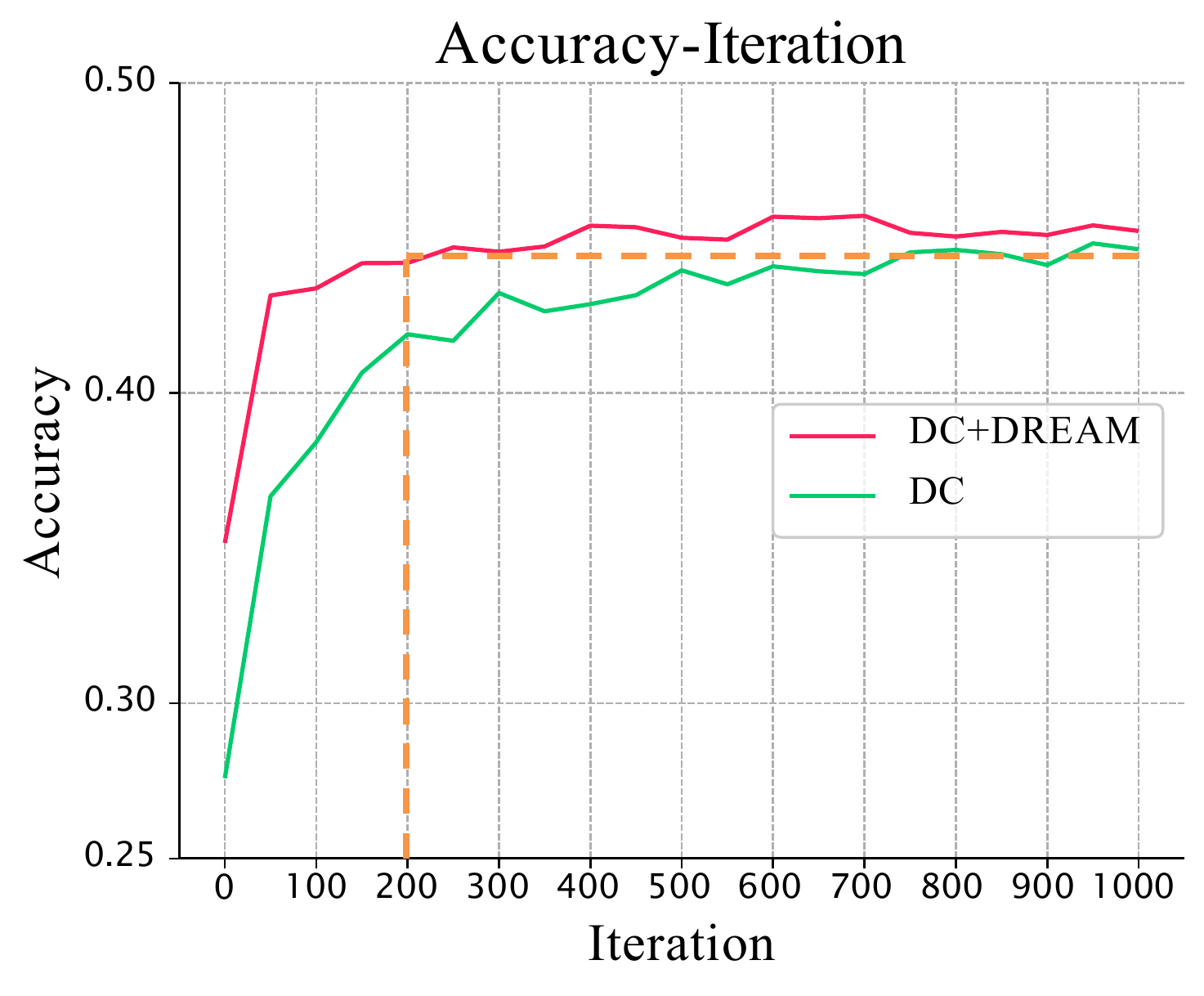}
    \caption{The accuracy curve of adding DREAM strategy to DC. }
\end{subfigure}
\begin{subfigure}{0.32\textwidth}
    \includegraphics[width=\textwidth]{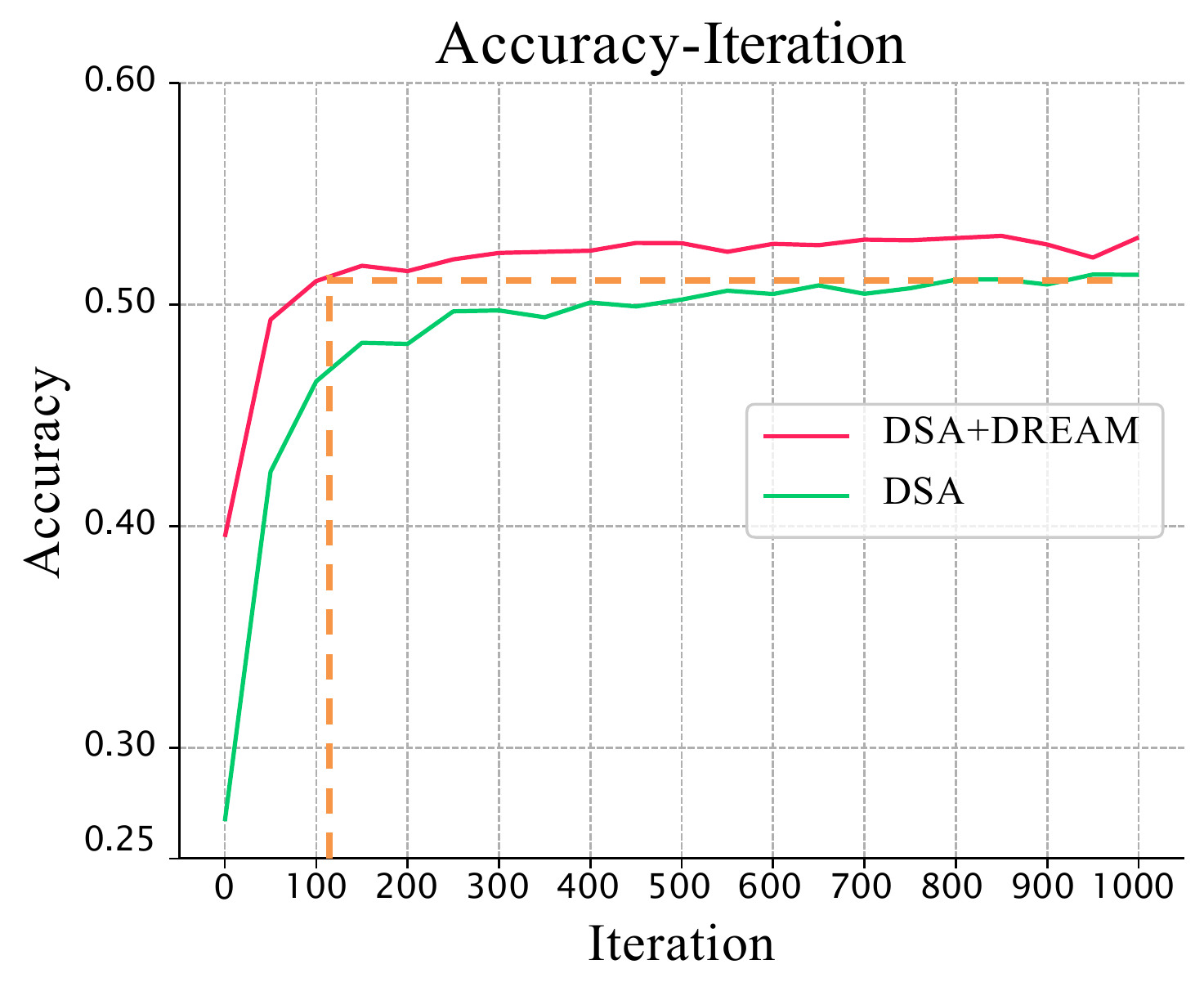}
    \caption{The accuracy curve of adding DREAM strategy to DSA. }
\end{subfigure}
\begin{subfigure}{0.32\textwidth}
    \includegraphics[width=\textwidth]{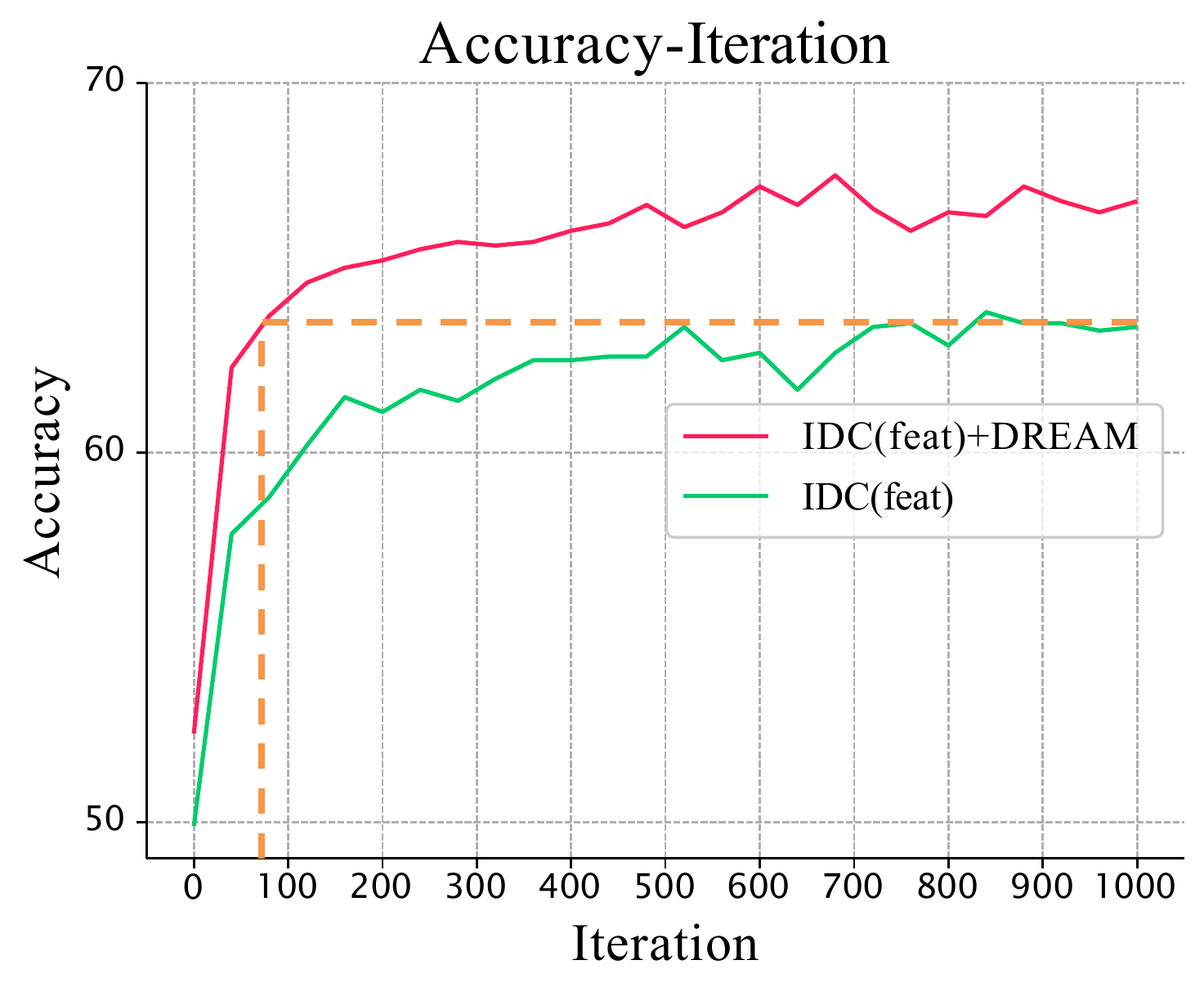}
    \caption{The accuracy curve of adding DREAM strategy to distribution matching. }
    \label{fig:curve-dm}
\end{subfigure}
\caption{Applying the DREAM strategy brings stable performance and efficiency improvements. }
\label{fig:curve}
\end{figure*}

In addition to gradient matching, we also explore the applicability of our method in embedding distribution matching. 
For distribution matching, the optimization is constrained by:
\begin{equation}
    \label{eqn:match}
    \mathcal{S}^*=\arg\min_\mathcal{S}\mathbf{D}\left(
    \xi(\mathcal{M}_\theta(\mathcal{A}(\mathcal{S}))),
    \xi(\mathcal{M}_\theta(\mathcal{A}(\mathcal{T})))\right),
\end{equation}
where $\xi$ represents averaging the features in the channel dimension. 
For gradient matching, the boundary samples generate large gradients which bias the optimization, while for distribution matching, the boundary samples shift the average feature. 
Random sampling introduces random factors to the shifts and degrades the matching efficiency. 
DREAM, on the contrary, ensures the evenness and diversity of the original images for matching. 
It largely reduces the feature shifts and consistently provide appropriate supervision for the optimization. 
We conduct the experiments on IDC~\cite{idc} with distribution matching. 
Under the setting of 10 images-per-class on the CIFAR10 dataset, the original IDC performs much poorer than gradient matching, which is also stated in~\cite{idc}. 
Applying DREAM completely reversed this situation by improving the dataset performance by a large margin. 
Besides, it only takes less than one tenth of the original iteration number to reach the performance, as shown in Fig.~\ref{fig:curve-dm}. 

\input{tables/other_methods.tex}
\input{tables/time.tex}

\section{Distilled Dataset Visualization}
\begin{figure*}[t]
\centering
\begin{subfigure}{0.33\textwidth}
    \includegraphics[width=\textwidth]{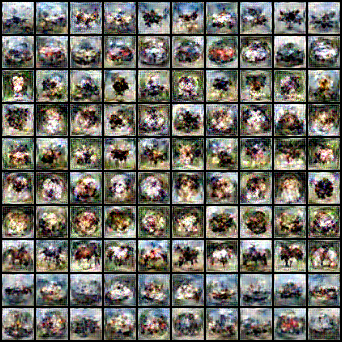}
    \caption{DC}
\end{subfigure}
\hskip 1em
\begin{subfigure}{0.33\textwidth}
    \includegraphics[width=\textwidth]{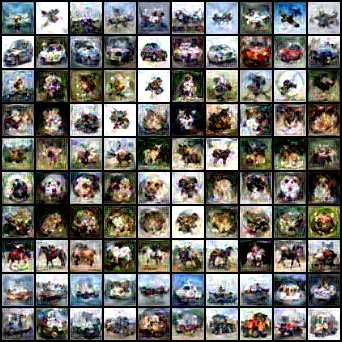}
    \caption{DC+DREAM}
\end{subfigure}
\begin{subfigure}{0.33\textwidth}
    \includegraphics[width=\textwidth]{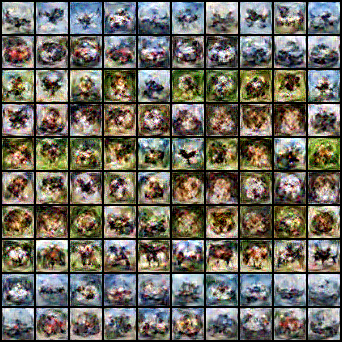}
    \caption{DSA}
\end{subfigure}
\hskip 1em
\begin{subfigure}{0.33\textwidth}
    \includegraphics[width=\textwidth]{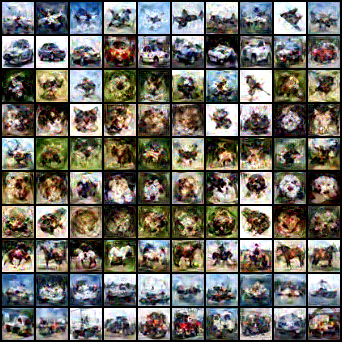}
    \caption{DSA+DREAM}
\end{subfigure}
\caption{Applying DREAM improves the image quality and sample diversity. }
\label{fig_supp:sample}
\end{figure*}

\begin{figure*}[t]
\centering
\begin{subfigure}{0.33\textwidth}
    \includegraphics[width=\textwidth]{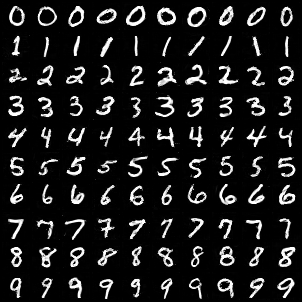}
    \caption{MNIST}
\end{subfigure}
\begin{subfigure}{0.33\textwidth}
    \includegraphics[width=\textwidth]{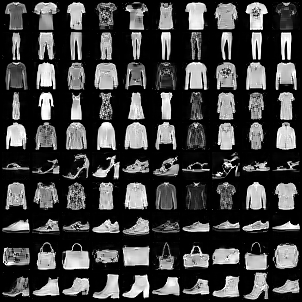}
    \caption{FashionMNIST}
\end{subfigure}
\begin{subfigure}{0.33\textwidth}
    \includegraphics[width=\textwidth]{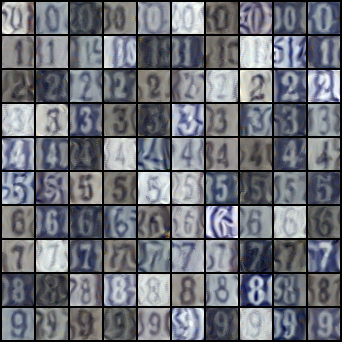}
    \caption{SVHN}
\end{subfigure}
\caption{Example visualizations of the distilled images on MNIST, FashionMNIST and SVHN. }
\label{fig:sample-mnist}
\end{figure*}

In order to more intuitively demonstrate the effects on the distilled images, we compare the distillation results of adding the proposed DREAM strategy or not in Fig.~\ref{fig_supp:sample}. 
DREAM improves the quality of the distilled datasets from two perspectives. 
Firstly, the images optimized by DREAM show more obvious categorical characteristics. 
Secondly, DREAM introduces more variety to the distilled images. 
With these two improvements, dream brings better performance to the distilled datasets. 

We provide some extra visualizations of the distilled images on MNIST, FashionMNIST and SVHN in Fig.~\ref{fig:sample-mnist}.

\section{Differences from Related Works}
There are some recent works focusing on improving the efficiency of dataset distillation. 
RFAD reduces the calculation of neural tangent kernel matrix in Kernel Inducing Points (KIP) from $O (|S^{2}|)$ to $O (|S|)$ by using random feature approximation~\cite{loo2022efficient}.
It takes into account the similarity between Neural Tangent Kernel (NTK) and Neural Network Gaussian Process (NNGP) kernels.
RFAD focuses on reducing the calculation complexity in KIP, while our proposed DREAM is aiming at improving the matching efficiency through selecting representative original images, which has no contradicts. 

Jiang et al. analyze the shortcomings of gradient matching method and propose the idea of matching multi-level gradients from the angle perspective~\cite{jiang2022delving}.
There are also many other methods~\cite{lorraine2020optimizing,vicol2022implicit}, which analyze the shortcomings of existing methods from the perspective of two-level optimization and improve the efficiency.
Comparatively, DREAM addresses the matching efficiency problem from the sampling perspective for both gradient matching and embedding distribution matching. 
DREAM is able to be easily plugged into other dataset distillation methods to significantly reduce the required training iterations. 

\section{Clustering Analysis}
We further analyze the extra time cost caused by the clustering process in Tab.~\ref{tab:time}.
For CIFAR-10, in each inner loop, the matching process and image updating cost 0.2s.
Every 10 inner loops, a clustering process is conducted. 
The clustering process takes 1s, so by average the clustering time for each inner loop is 0.1s. 
The total average inner loop time is 0.3s, compared to the original 0.2s.
Considering that we only need one tenth to one fifth of the iterations to obtain the original performance, we save more than 70$\%$ of the time.
For CIFAR-100 with more classes, the extra clustering time is one twentieth of the original image updating time, which is negligible. DREAM significantly improves the matching efficiency and reduces the required training time for dataset distillation.

%% file: tables/other_methods.tex
\begin{table*}[t]
\caption{Top-1 accuracy of test models trained on distilled synthetic images on multiple datasets. The distillation training is conducted with ConvNet-3. }
\label{tab:other_methods}
\centering
\small
\setlength{\tabcolsep}{8pt}
\begin{tabular}{c|ccc|ccc|ccc|ccc}
\toprule
\multirow{2}{*}{Dataset}&
\multicolumn{3}{c|}{MNIST}   & \multicolumn{3}{c|}{SVHN} & \multicolumn{3}{c|}{CIFAR10} & \multicolumn{3}{c}{CIFAR100}\\
 & 1 & 10 & 50 & 1 & 10 & 50 & 1 & 10 & 50 & 1 & 10 & 50  \\ \midrule
DD~\cite{wang2018dataset} & - & 79.5 & - & - & - & - & - & 36.8 & - & - & - & -\\
LD~\cite{LD} & 60.9 & 87.3 & 93.3 & - & - & - & 25.7 & 38.3 & 42.5 & 11.5 & - & - \\
DC~\cite{zhao2020dc} & 91.7 & 97.4 & 98.8 & 31.2 & 76.1 & 82.3 & 28.3 & 44.9 & 53.9 & 12.8 & 25.2 & - \\
DSA~\cite{zhao2021dsa} & 88.7 & 97.8 & 99.2 & 27.5 & 79.2 & 84.4 & 28.8 & 52.1 & 60.6 & 13.9 & 32.3 & 42.8 \\
DM~\cite{dm} & 89.7 & 97.5 & 98.6 & - & - & - & 26.0 & 48.9 & 63.0 & 11.4 & 29.7 & 43.6 \\
CAFE~\cite{wang2022cafe} & 93.1 & 97.2 & 98.6 & 42.6 & 75.9 & 81.3 & 30.3 & 46.3 & 55.5 & 12.9 & 27.8 & 37.9 \\
MTT~\cite{mtt} & - & - & - & - & - & - & 46.3 & 65.3 & 71.6 & 24.3 & 40.1 & 47.7 \\
IDC~\cite{idc} & 94.2 & 98.4 & 99.1 & 68.5 & 87.5 & 90.1 & 50.6 & 67.5 & 74.5 & - & 45.1 & - \\
KIP~\cite{nguyen2020dataset,nguyen2021dataset} & 90.1 & 97.5 & 98.3 & 57.3 & 75.0 & 80.5 & 49.9 & 62.7 & 68.6 & 15.7 & 28.3 & -  \\ 
RFAD~\cite{loo2022efficient} & 94.4 & 98.5 & 98.8 & 52.2 & 74.9 & 80.9 & \textbf{53.6} & 66.3 & 71.1 & 26.3 & 33.0 & -  \\
HaBa~\cite{liu2022dataset} & 92.4 & 97.4 & 98.1 & 69.8 & 83.2 & 88.3 & 48.3 & \textbf{69.9} & 74.0 & \textbf{33.4} & 40.2 & 47.0  \\
FRePo~\cite{zhou2022dataset} & 93.0 & 98.6 & 99.2 & - & - & - & 46.8 & 65.5 & 71.7 & 28.7 & 42.5 & 44.3  \\
DREAM & \textbf{95.7} & \textbf{98.6} & \textbf{99.2} & \textbf{69.8} & \textbf{87.9} & \textbf{90.5} & 51.1 & 69.4 & \textbf{74.8} & 29.5 & \textbf{46.8} & \textbf{52.6} \\ \bottomrule
\end{tabular}
\end{table*}

%% file: tables/time.tex
\begin{table}[t]
    \caption{Time cost of adding DREAM strategy (s).}
    \label{tab:time}
    \centering
    \small
    \begin{tabular}{lcccc}
    \toprule
        \multirow{2}{*}{Datasets} & \multirow{2}{*}{Methods} & \multirow{2}{*}{Clustering} & Update & Inner \\
        & & & Images & Loop \\
        \midrule
        \multirow{2}{*}{CIFAR-10}& IDC~\cite{idc} & - & 0.2 & 0.2 \\ 
        & DREAM & 0.1 & 0.2 & 0.3 \\
        \midrule
        \multirow{2}{*}{CIFAR-100}& IDC~\cite{idc} & - & 2.0 & 2.0 \\ 
        & DREAM & 0.1 & 2.0 & 2.1 \\
        \bottomrule
    \end{tabular}
\end{table}

%% file: egpaper_final.bbl
\begin{thebibliography}{10}\itemsep=-1pt

\bibitem{aljundi2019gradient}
Rahaf Aljundi, Min Lin, Baptiste Goujaud, and Yoshua Bengio.
\newblock Gradient based sample selection for online continual learning.
\newblock In {\em NeurIPS}, pages 11817--11826, 2019.

\bibitem{k-means++}
David Arthur and Sergei Vassilvitskii.
\newblock k-means++: The advantages of careful seeding.
\newblock Technical report, Stanford, 2006.

\bibitem{LD}
Ondrej Bohdal, Yongxin Yang, and Timothy Hospedales.
\newblock Flexible dataset distillation: Learn labels instead of images.
\newblock {\em arXiv preprint arXiv:2006.08572}, 2020.

\bibitem{mtt}
George Cazenavette, Tongzhou Wang, Antonio Torralba, Alexei~A Efros, and
  Jun-Yan Zhu.
\newblock Dataset distillation by matching training trajectories.
\newblock In {\em CVPR}, pages 4750--4759, 2022.

\bibitem{coleman2019selection}
Cody Coleman, Christopher Yeh, Stephen Mussmann, Baharan Mirzasoleiman, Peter
  Bailis, Percy Liang, Jure Leskovec, and Matei Zaharia.
\newblock Selection via proxy: Efficient data selection for deep learning.
\newblock {\em arXiv preprint arXiv:1906.11829}, 2019.

\bibitem{cui2022dc}
Justin Cui, Ruochen Wang, Si Si, and Cho-Jui Hsieh.
\newblock Dc-bench: Dataset condensation benchmark.
\newblock In {\em NeurIPS}, 2022.

\bibitem{cui2023scaling}
Justin Cui, Ruochen Wang, Si Si, and Cho-Jui Hsieh.
\newblock Scaling up dataset distillation to imagenet-1k with constant memory.
\newblock In {\em International Conference on Machine Learning}, pages
  6565--6590. PMLR, 2023.

\bibitem{danelljan2017eco}
Martin Danelljan, Goutam Bhat, Fahad Shahbaz~Khan, and Michael Felsberg.
\newblock Eco: Efficient convolution operators for tracking.
\newblock In {\em CVPR}, pages 6638--6646, 2017.

\bibitem{zhou2023dataset}
Zhou Daquan, Kai Wang, Jianyang Gu, Xiangyu Peng, Dongze Lian, Yifan Zhang,
  Yang You, and Jiashi Feng.
\newblock Dataset quantization.
\newblock In {\em Proceedings of the IEEE/CVF International Conference on
  Computer Vision}, 2023.

\bibitem{deng2009imagenet}
Jia Deng, Wei Dong, Richard Socher, Li-Jia Li, Kai Li, and Li Fei-Fei.
\newblock Imagenet: A large-scale hierarchical image database.
\newblock In {\em CVPR}, pages 248--255. Ieee, 2009.

\bibitem{hierarchical}
Chris Ding and Xiaofeng He.
\newblock Cluster merging and splitting in hierarchical clustering algorithms.
\newblock In {\em ICDM.}, pages 139--146. IEEE, 2002.

\bibitem{dosovitskiy2020image}
Alexey Dosovitskiy, Lucas Beyer, Alexander Kolesnikov, Dirk Weissenborn,
  Xiaohua Zhai, Thomas Unterthiner, Mostafa Dehghani, Matthias Minderer, Georg
  Heigold, Sylvain Gelly, et~al.
\newblock An image is worth 16x16 words: Transformers for image recognition at
  scale.
\newblock {\em arXiv preprint arXiv:2010.11929}, 2020.

\bibitem{ftd}
Jiawei Du, Yidi Jiang, Vincent Y.~F. Tan, Joey~Tianyi Zhou, and Haizhou Li.
\newblock Minimizing the accumulated trajectory error to improve dataset
  distillation.
\newblock In {\em CVPR}, pages 3749--3758, 2023.

\bibitem{dbscan}
Martin Ester, Hans-Peter Kriegel, J{\"o}rg Sander, Xiaowei Xu, et~al.
\newblock A density-based algorithm for discovering clusters in large spatial
  databases with noise.
\newblock In {\em KDD}, pages 226--231, 1996.

\bibitem{k-means}
Edward~W Forgy.
\newblock Cluster analysis of multivariate data: efficiency versus
  interpretability of classifications.
\newblock {\em Biometrics}, 21:768--769, 1965.

\bibitem{gidaris2018dynamic}
Spyros Gidaris and Nikos Komodakis.
\newblock Dynamic few-shot visual learning without forgetting.
\newblock In {\em CVPR}, pages 4367--4375, 2018.

\bibitem{goodfellow2020generative}
Ian Goodfellow, Jean Pouget-Abadie, Mehdi Mirza, Bing Xu, David Warde-Farley,
  Sherjil Ozair, Aaron Courville, and Yoshua Bengio.
\newblock Generative adversarial networks.
\newblock {\em Communications of the ACM}, 63(11):139--144, 2020.

\bibitem{guo2022deepcore}
Chengcheng Guo, Bo Zhao, and Yanbing Bai.
\newblock Deepcore: A comprehensive library for coreset selection in deep
  learning.
\newblock {\em arXiv preprint arXiv:2204.08499}, 2022.

\bibitem{intra-distance}
Greg Hamerly and Charles Elkan.
\newblock Alternatives to the k-means algorithm that find better clusterings.
\newblock In {\em CIKM}, pages 600--607, 2002.

\bibitem{he2022masked}
Kaiming He, Xinlei Chen, Saining Xie, Yanghao Li, Piotr Doll{\'a}r, and Ross
  Girshick.
\newblock Masked autoencoders are scalable vision learners.
\newblock In {\em CVPR}, pages 16000--16009, 2022.

\bibitem{he2016deep}
Kaiming He, Xiangyu Zhang, Shaoqing Ren, and Jian Sun.
\newblock Deep residual learning for image recognition.
\newblock In {\em CVPR}, pages 770--778, 2016.

\bibitem{huang2017densely}
Gao Huang, Zhuang Liu, Laurens Van Der~Maaten, and Kilian~Q Weinberger.
\newblock Densely connected convolutional networks.
\newblock In {\em CVPR}, pages 4700--4708, 2017.

\bibitem{jiang2022delving}
Zixuan Jiang, Jiaqi Gu, Mingjie Liu, and David~Z Pan.
\newblock Delving into effective gradient matching for dataset condensation.
\newblock {\em arXiv preprint arXiv:2208.00311}, 2022.

\bibitem{karras2020training}
Tero Karras, Miika Aittala, Janne Hellsten, Samuli Laine, Jaakko Lehtinen, and
  Timo Aila.
\newblock Training generative adversarial networks with limited data.
\newblock {\em NeurIPS}, 33:12104--12114, 2020.

\bibitem{idc}
Jang-Hyun Kim, Jinuk Kim, Seong~Joon Oh, Sangdoo Yun, Hwanjun Song, Joonhyun
  Jeong, Jung-Woo Ha, and Hyun~Oh Song.
\newblock Dataset condensation via efficient synthetic-data parameterization.
\newblock {\em arXiv preprint arXiv:2205.14959}, 2022.

\bibitem{krizhevsky2009learning}
Alex Krizhevsky, Geoffrey Hinton, et~al.
\newblock Learning multiple layers of features from tiny images.
\newblock 2009.

\bibitem{lapedriza2013all}
Agata Lapedriza, Hamed Pirsiavash, Zoya Bylinskii, and Antonio Torralba.
\newblock Are all training examples equally valuable?
\newblock {\em arXiv preprint arXiv:1311.6510}, 2013.

\bibitem{lecun1998gradient}
Yann LeCun, L{\'e}on Bottou, Yoshua Bengio, and Patrick Haffner.
\newblock Gradient-based learning applied to document recognition.
\newblock {\em Proceedings of the IEEE}, 86(11):2278--2324, 1998.

\bibitem{liu2022dataset}
Songhua Liu, Kai Wang, Xingyi Yang, Jingwen Ye, and Xinchao Wang.
\newblock Dataset distillation via factorization.
\newblock In {\em NeurIPS}, 2022.

\bibitem{liu2021swin}
Ze Liu, Yutong Lin, Yue Cao, Han Hu, Yixuan Wei, Zheng Zhang, Stephen Lin, and
  Baining Guo.
\newblock Swin transformer: Hierarchical vision transformer using shifted
  windows.
\newblock In {\em ICCV}, pages 10012--10022, 2021.

\bibitem{loo2022efficient}
Noel Loo, Ramin Hasani, Alexander Amini, and Daniela Rus.
\newblock Efficient dataset distillation using random feature approximation.
\newblock In {\em NeurIPS}, 2022.

\bibitem{lorraine2020optimizing}
Jonathan Lorraine, Paul Vicol, and David Duvenaud.
\newblock Optimizing millions of hyperparameters by implicit differentiation.
\newblock In {\em International Conference on Artificial Intelligence and
  Statistics}, pages 1540--1552. PMLR, 2020.

\bibitem{netzer2011reading}
Yuval Netzer, Tao Wang, Adam Coates, Alessandro Bissacco, Bo Wu, and Andrew~Y
  Ng.
\newblock Reading digits in natural images with unsupervised feature learning.
\newblock 2011.

\bibitem{nguyen2020dataset}
Timothy Nguyen, Zhourong Chen, and Jaehoon Lee.
\newblock Dataset meta-learning from kernel ridge-regression.
\newblock In {\em ICLR}, 2020.

\bibitem{nguyen2021dataset}
Timothy Nguyen, Roman Novak, Lechao Xiao, and Jaehoon Lee.
\newblock Dataset distillation with infinitely wide convolutional networks.
\newblock {\em NeurIPS}, 34:5186--5198, 2021.

\bibitem{Omer_fast-pytorch-kmeans_2020}
Sehban Omer.
\newblock {fast-pytorch-kmeans}, 9 2020.

\bibitem{qin2023infobatch}
Ziheng Qin, Kai Wang, Zangwei Zheng, Jianyang Gu, Xiangyu Peng, Daquan Zhou,
  and Yang You.
\newblock Infobatch: Lossless training speed up by unbiased dynamic data
  pruning.
\newblock {\em arXiv preprint arXiv:2303.04947}, 2023.

\bibitem{rebuffi2017icarl}
Sylvestre-Alvise Rebuffi, Alexander Kolesnikov, Georg Sperl, and Christoph~H
  Lampert.
\newblock icarl: Incremental classifier and representation learning.
\newblock In {\em CVPR}, pages 2001--2010, 2017.

\bibitem{redmon2016you}
Joseph Redmon, Santosh Divvala, Ross Girshick, and Ali Farhadi.
\newblock You only look once: Unified, real-time object detection.
\newblock In {\em CVPR}, pages 779--788, 2016.

\bibitem{rehioui2016denclue}
Hajar Rehioui, Abdellah Idrissi, Manar Abourezq, and Faouzia Zegrari.
\newblock Denclue-im: A new approach for big data clustering.
\newblock {\em Procedia Computer Science}, 83:560--567, 2016.

\bibitem{sener2017active}
Ozan Sener and Silvio Savarese.
\newblock Active learning for convolutional neural networks: A core-set
  approach.
\newblock {\em arXiv preprint arXiv:1708.00489}, 2017.

\bibitem{shleifer2019using}
Sam Shleifer and Eric Prokop.
\newblock Using small proxy datasets to accelerate hyperparameter search.
\newblock {\em arXiv preprint arXiv:1906.04887}, 2019.

\bibitem{sorscher2022beyond}
Ben Sorscher, Robert Geirhos, Shashank Shekhar, Surya Ganguli, and Ari Morcos.
\newblock Beyond neural scaling laws: beating power law scaling via data
  pruning.
\newblock {\em Advances in Neural Information Processing Systems},
  35:19523--19536, 2022.

\bibitem{toneva2018empirical}
Mariya Toneva, Alessandro Sordoni, Remi Tachet~des Combes, Adam Trischler,
  Yoshua Bengio, and Geoffrey~J Gordon.
\newblock An empirical study of example forgetting during deep neural network
  learning.
\newblock {\em arXiv preprint arXiv:1812.05159}, 2018.

\bibitem{tran2020towards}
Ngoc-Trung Tran, Viet-Hung Tran, Ngoc-Bao Nguyen, Trung-Kien Nguyen, and
  Ngai-Man Cheung.
\newblock Towards good practices for data augmentation in gan training.
\newblock {\em arXiv preprint arXiv:2006.05338}, 2:3, 2020.

\bibitem{tsai2018learning}
Yi-Hsuan Tsai, Wei-Chih Hung, Samuel Schulter, Kihyuk Sohn, Ming-Hsuan Yang,
  and Manmohan Chandraker.
\newblock Learning to adapt structured output space for semantic segmentation.
\newblock In {\em CVPR}, pages 7472--7481, 2018.

\bibitem{ulyanov2016instance}
Dmitry Ulyanov, Andrea Vedaldi, and Victor Lempitsky.
\newblock Instance normalization: The missing ingredient for fast stylization.
\newblock {\em arXiv preprint arXiv:1607.08022}, 2016.

\bibitem{vicol2022implicit}
Paul Vicol, Jonathan~P Lorraine, Fabian Pedregosa, David Duvenaud, and Roger~B
  Grosse.
\newblock On implicit bias in overparameterized bilevel optimization.
\newblock In {\em International Conference on Machine Learning}, pages
  22234--22259. PMLR, 2022.

\bibitem{wang2023dim}
Kai Wang, Jianyang Gu, Daquan Zhou, Zheng Zhu, Wei Jiang, and Yang You.
\newblock Dim: Distilling dataset into generative model.
\newblock {\em arXiv preprint arXiv:2303.04707}, 2023.

\bibitem{wang2022cafe}
Kai Wang, Bo Zhao, Xiangyu Peng, Zheng Zhu, Shuo Yang, Shuo Wang, Guan Huang,
  Hakan Bilen, Xinchao Wang, and Yang You.
\newblock Cafe: Learning to condense dataset by aligning features.
\newblock In {\em CVPR}, pages 12196--12205, 2022.

\bibitem{wang2018dataset}
Tongzhou Wang, Jun-Yan Zhu, Antonio Torralba, and Alexei~A Efros.
\newblock Dataset distillation.
\newblock {\em arXiv preprint arXiv:1811.10959}, 2018.

\bibitem{wiewel2021condensed}
Felix Wiewel and Bin Yang.
\newblock Condensed composite memory continual learning.
\newblock In {\em IJCNN}, pages 1--8. IEEE, 2021.

\bibitem{xiao2017fashion}
Han Xiao, Kashif Rasul, and Roland Vollgraf.
\newblock Fashion-mnist: a novel image dataset for benchmarking machine
  learning algorithms.
\newblock {\em arXiv preprint arXiv:1708.07747}, 2017.

\bibitem{zhao2021dsa}
Bo Zhao and Hakan Bilen.
\newblock Dataset condensation with differentiable siamese augmentation.
\newblock In {\em ICML}, pages 12674--12685. PMLR, 2021.

\bibitem{dm}
Bo Zhao and Hakan Bilen.
\newblock Dataset condensation with distribution matching.
\newblock {\em arXiv preprint arXiv:2110.04181}, 2021.

\bibitem{zhao2020dc}
Bo Zhao, Konda~Reddy Mopuri, and Hakan Bilen.
\newblock Dataset condensation with gradient matching.
\newblock In {\em ICLR}, 2020.

\bibitem{zhao2020differentiable}
Shengyu Zhao, Zhijian Liu, Ji Lin, Jun-Yan Zhu, and Song Han.
\newblock Differentiable augmentation for data-efficient gan training.
\newblock {\em NeurIPS}, 33:7559--7570, 2020.

\bibitem{zhao2020image}
Zhengli Zhao, Zizhao Zhang, Ting Chen, Sameer Singh, and Han Zhang.
\newblock Image augmentations for gan training.
\newblock {\em arXiv preprint arXiv:2006.02595}, 2020.

\bibitem{zheng2023preventing}
Zangwei Zheng, Mingyuan Ma, Kai Wang, Ziheng Qin, Xiangyu Yue, and Yang You.
\newblock Preventing zero-shot transfer degradation in continual learning of
  vision-language models.
\newblock {\em arXiv preprint arXiv:2303.06628}, 2023.

\bibitem{zhou2022dataset}
Yongchao Zhou, Ehsan Nezhadarya, and Jimmy Ba.
\newblock Dataset distillation using neural feature regression.
\newblock In {\em NeurIPS}, 2022.

\end{thebibliography}
